\documentclass{egpubl}
\usepackage{pg2022}

\SpecialIssuePaper         

\usepackage[T1]{fontenc}
\usepackage{dfadobe}  

\usepackage{cite}  
\BibtexOrBiblatex
\electronicVersion
\PrintedOrElectronic
\ifpdf \usepackage[pdftex]{graphicx} \pdfcompresslevel=9
\else \usepackage[dvips]{graphicx} \fi

\usepackage{egweblnk} 
\usepackage{diagbox}
\usepackage{longtable}
\usepackage{threeparttable}
\usepackage{epsfig}
\usepackage{subfigure}
\usepackage{paralist}
\usepackage{graphicx}
\usepackage{amsmath,amssymb} 
\usepackage{pgfplots}
\usepackage{tikz}
\usepackage{color}
\usepackage{xcolor}
\usepackage{booktabs,siunitx} 
\usepackage{makecell}
\usepackage{multirow}
\usepackage{array}

\title[Semi-MoreGAN: Semi-supervised Generative Adversarial Network for Mixture of Rain Removal]%
      {Semi-MoreGAN: Semi-supervised Generative Adversarial Network for Mixture of Rain Removal}

\author[Yiyang Shen et al.]
{\parbox{\textwidth}{\centering 
        Yiyang Shen$^{1}$\orcid{0000-0002-5468-3383}, 
        Yongzhen Wang$^{1}$,
         Mingqiang Wei$^{1\dag}$,
        Honghua Chen$^{1}$,
        Haoran Xie$^{2}$,
        Gary Cheng$^{3}$\thanks{Co-corresponding authors: M. Wei and G. Cheng.},
         Fu Lee Wang$^{4}$
        }
        \\
{\parbox{\textwidth}{\centering $^1$School of Computer Science and Technology, Nanjing University of Aeronautics and Astronautics, Nanjing, China\\
         $^2$School of Computing and Decision Sciences, Lingnan University, Hong Kong, China\\
         $^3$School of Mathematics and Information Technology, Education University of Hong Kong, Hong Kong, China\\
         $^4$School of Science and Technology, Hong Kong Metropolitan University, Hong Kong, China
       }
}
}

\begin{document}


\maketitle
\begin{abstract}
Real-world rain is a mixture of \textit{rain streaks} and \textit{rainy haze}.
However, current efforts formulate image rain streaks removal and rainy haze removal as separated models, worsening the loss of image details.
This paper attempts to solve the mixture of rain removal problem in a single model by estimating the scene depths of images. 
To this end, we propose a novel \textbf{SEMI}-supervised \textbf{M}ixture \textbf{O}f rain \textbf{RE}moval \textbf{G}enerative \textbf{A}dversarial \textbf{N}etwork (Semi-MoreGAN).
Unlike most of existing methods, Semi-MoreGAN is a joint learning paradigm of mixture of rain removal and depth estimation; and it effectively integrates the image features with the depth information for better rain removal. Furthermore, it leverages unpaired real-world rainy and clean images to bridge the gap between synthetic and real-world rain.
Extensive experiments show clear improvements of our approach over twenty representative state-of-the-arts on both synthetic and real-world rainy images. Source code is available at \textcolor{magenta}{ \href{https://github.com/syy-whu/Semi-MoreGAN}{https://github.com/syy-whu/Semi-MoreGAN}}.


\end{abstract}  
\section{Introduction}

Images captured on rainy days inevitably suffer from noticeable visibility degradations, e.g., content obstruction, color distortion, and detail blurring. Such degradations dramatically impair the visual perception quality and interfere with the performance of various outdoor vision systems such as intelligent surveillance~\cite{buch2011review}, autonomous driving~\cite{janai2020computer}, and object tracking~\cite{hu2018sinet}. Consequently, recovering clean images from their rain-polluted versions (called image rain removal) has attracted increasing attention in vision communities.

Image rain removal is a typical ill-posed problem. To make it well-posed, conventional wisdom explores various hand-crafted priors, such as Gaussian Mixture Model~\cite{li2016rain}, sparse coding~\cite{p_kang2011automatic} and low-rank representations~\cite{p_du2018single}. Although improving the overall visibility, these prior-based approaches cannot perform well with the complex rain distributions, e.g., varied angles, locations, depths, intensities, etc. Recently, deep neural networks~\cite{jiang2020multi, yang2017deep, wang2019spatial} have shown their effectiveness in coping with this problem. By learning a complex model from massive rainy/clean image pairs, the performance of these approaches is substantially improved. 
\begin{figure*}[!ht] \centering
\centering
\subfigure[Rainy image]{
\includegraphics[width=0.24\linewidth]{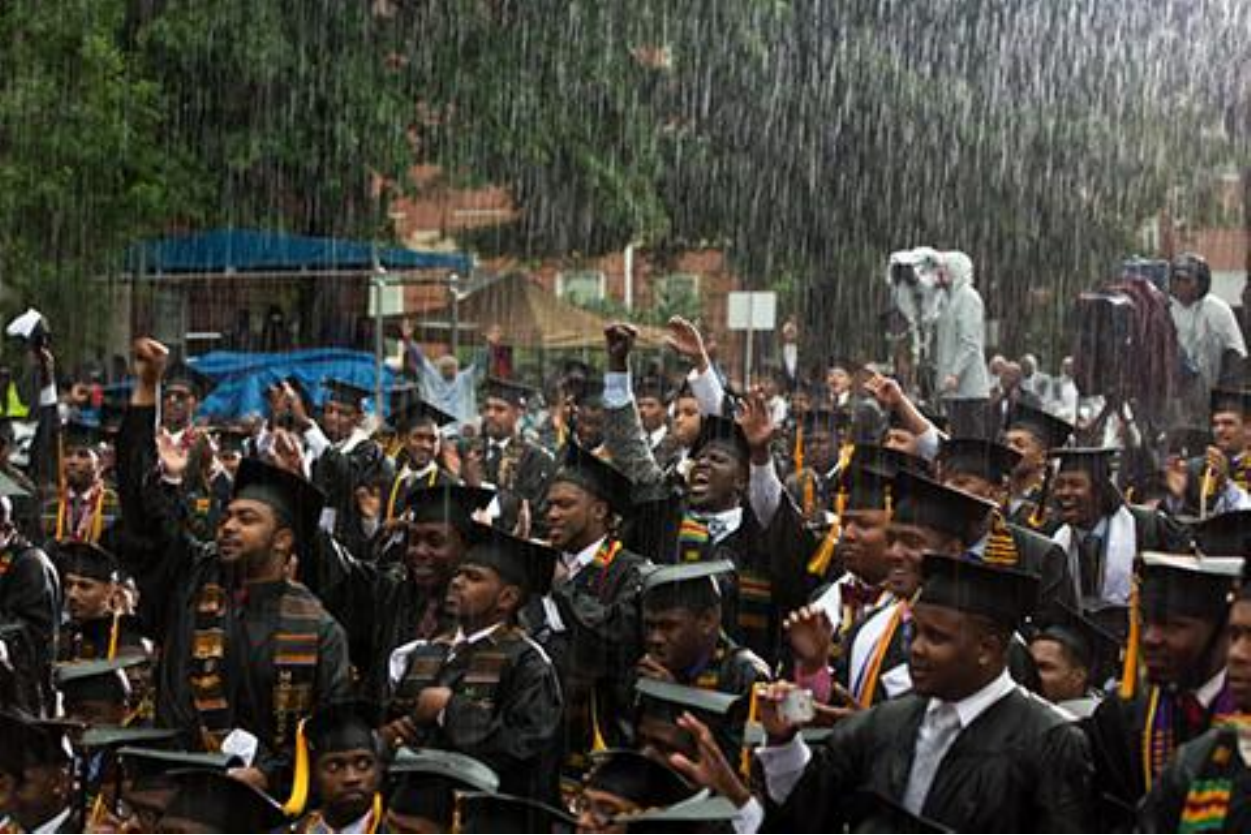}
}\hspace{-2mm}
\subfigure[MSPFN \cite{jiang2020multi}]{
\includegraphics[width=0.24\linewidth]{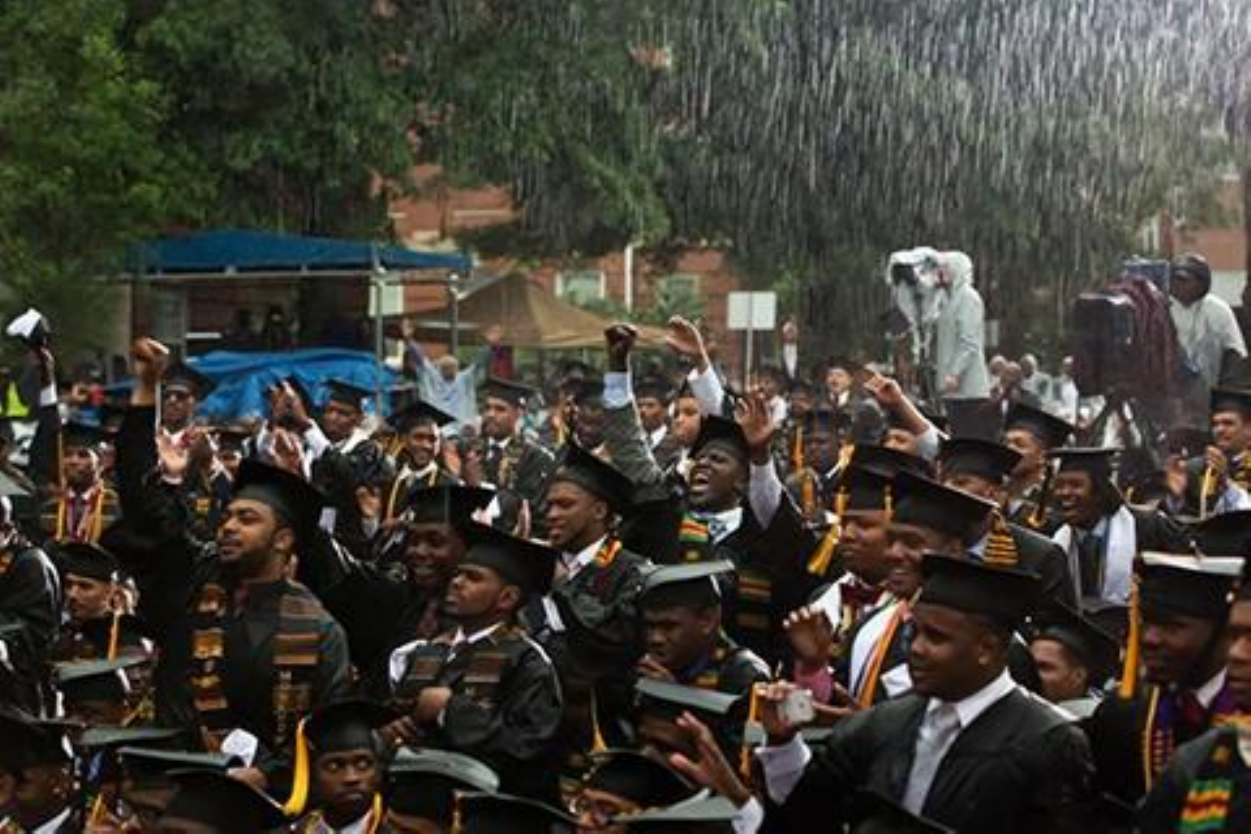}
}\hspace{-2mm}
\subfigure[MPRNet \cite{zamir2021multi}]{
\includegraphics[width=0.24\linewidth]{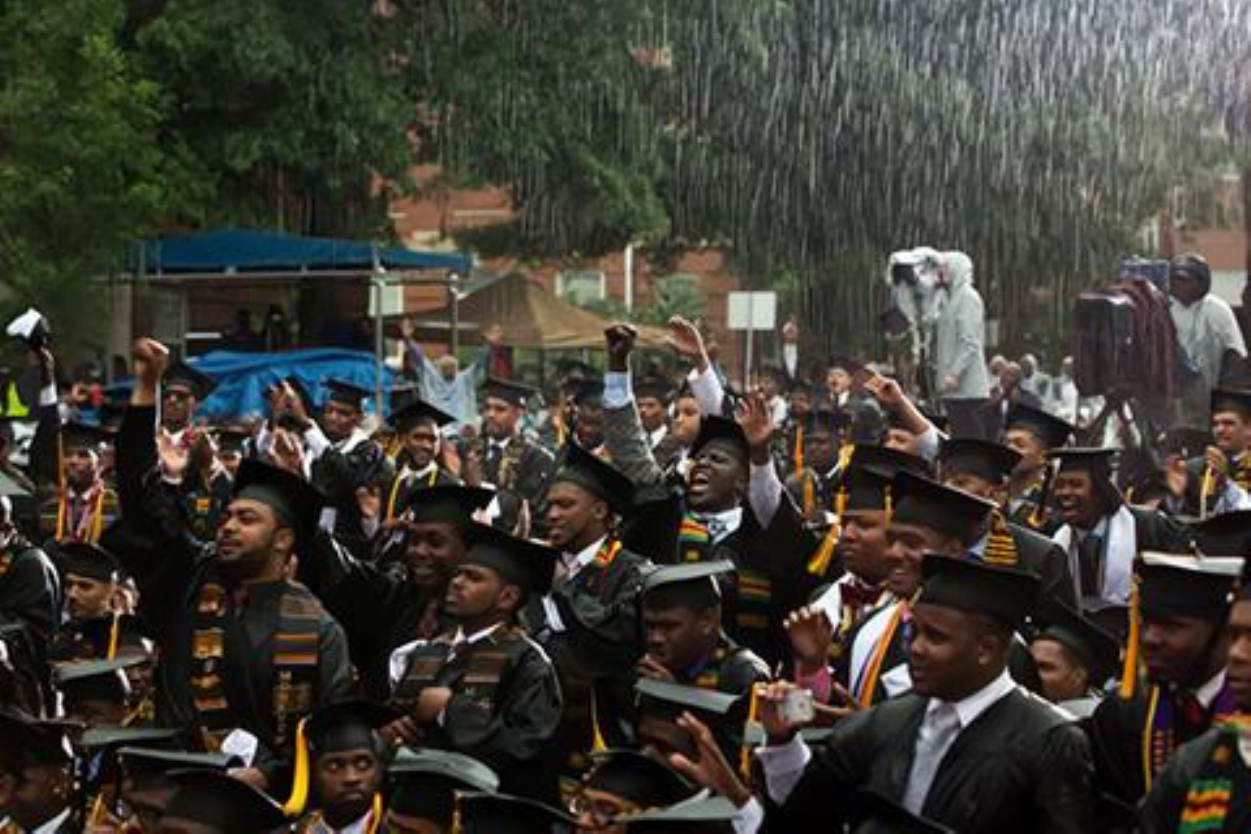}
}\hspace{-2mm}
\subfigure[Syn2Real \cite{yasarla2020syn2real}]{
\includegraphics[width=0.24\linewidth]{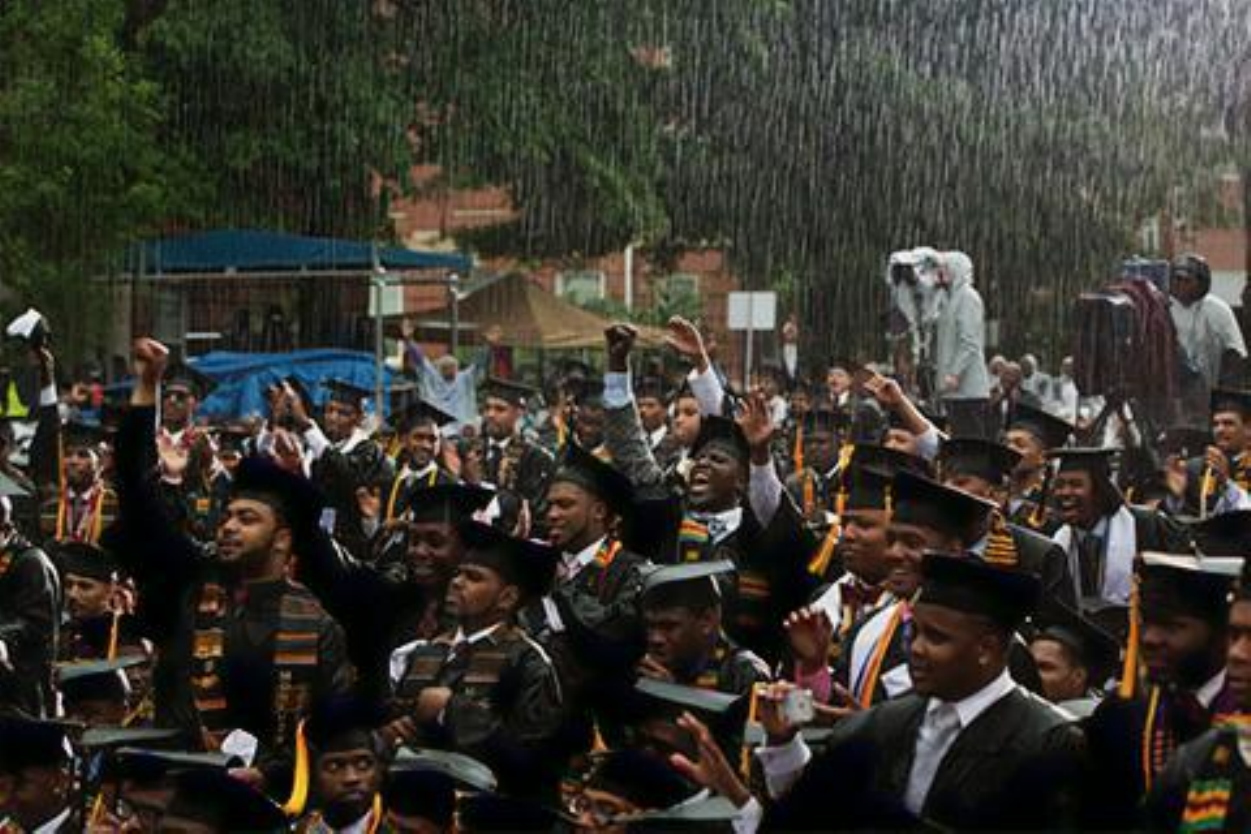}
}
\subfigure[JRGR \cite{ye2021closing}]{
\includegraphics[width=0.24\linewidth]{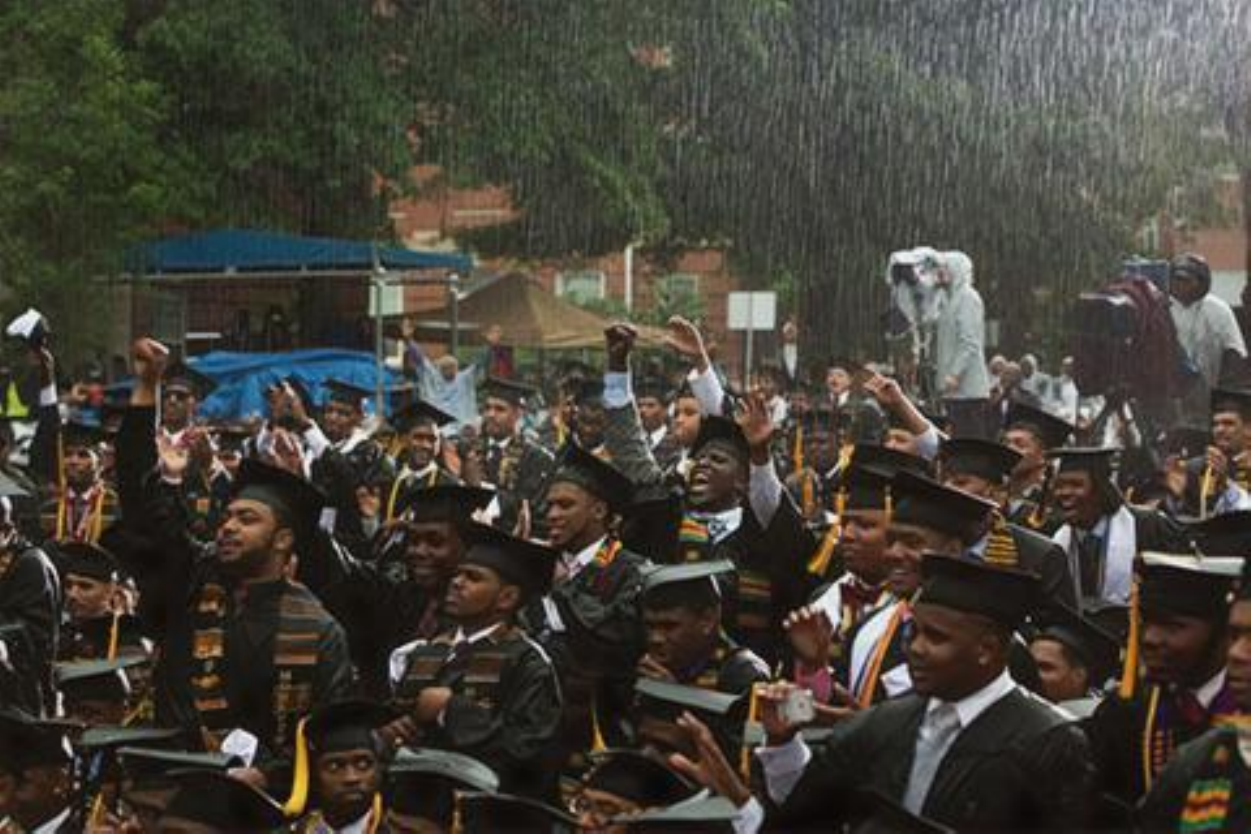}
}\hspace{-2mm}
\subfigure[SPANet \cite{wang2019spatial}+FFA \cite{qin2020ffa}]{
\includegraphics[width=0.24\linewidth]{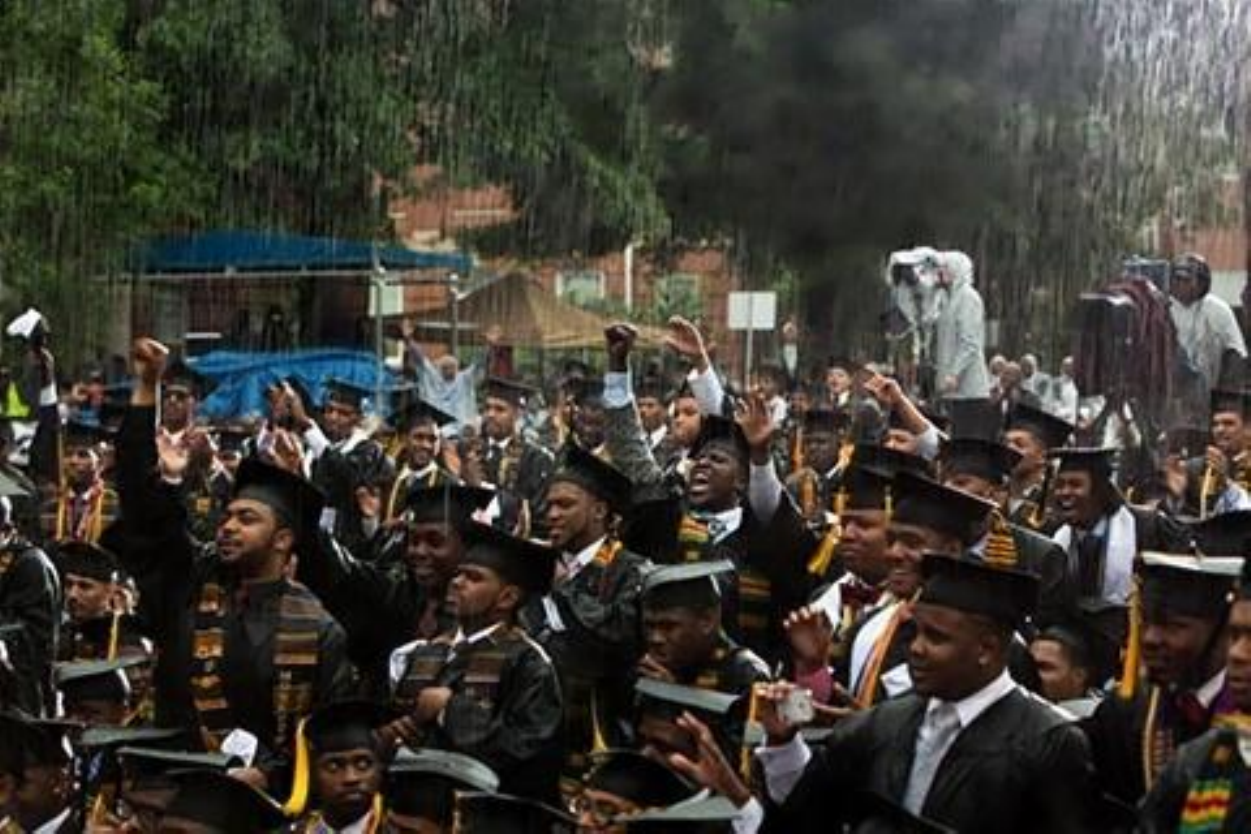}
}\hspace{-2mm}
\subfigure[DGNL-Net \cite{hu2021single}]{
\includegraphics[width=0.24\linewidth]{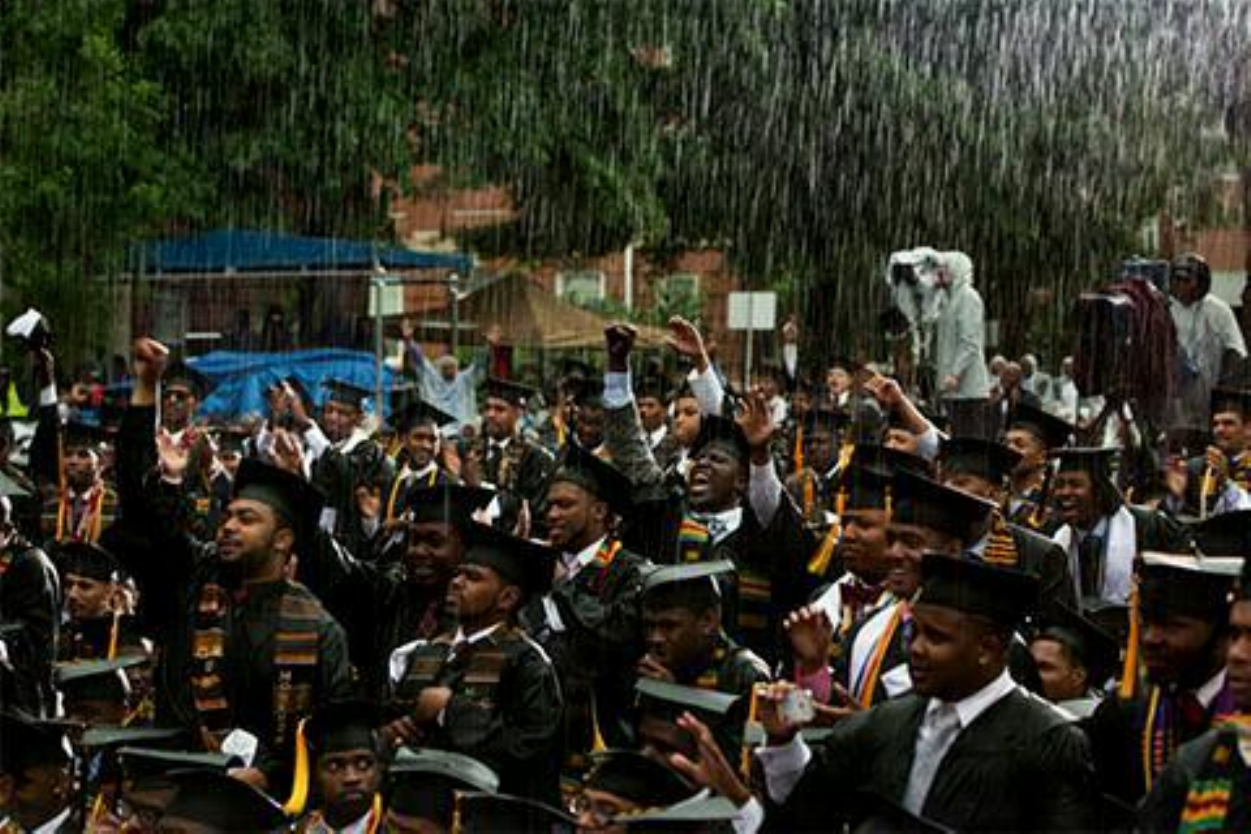}
}\hspace{-2mm}
\subfigure[Semi-MoreGAN]{
\includegraphics[width=0.24\linewidth]{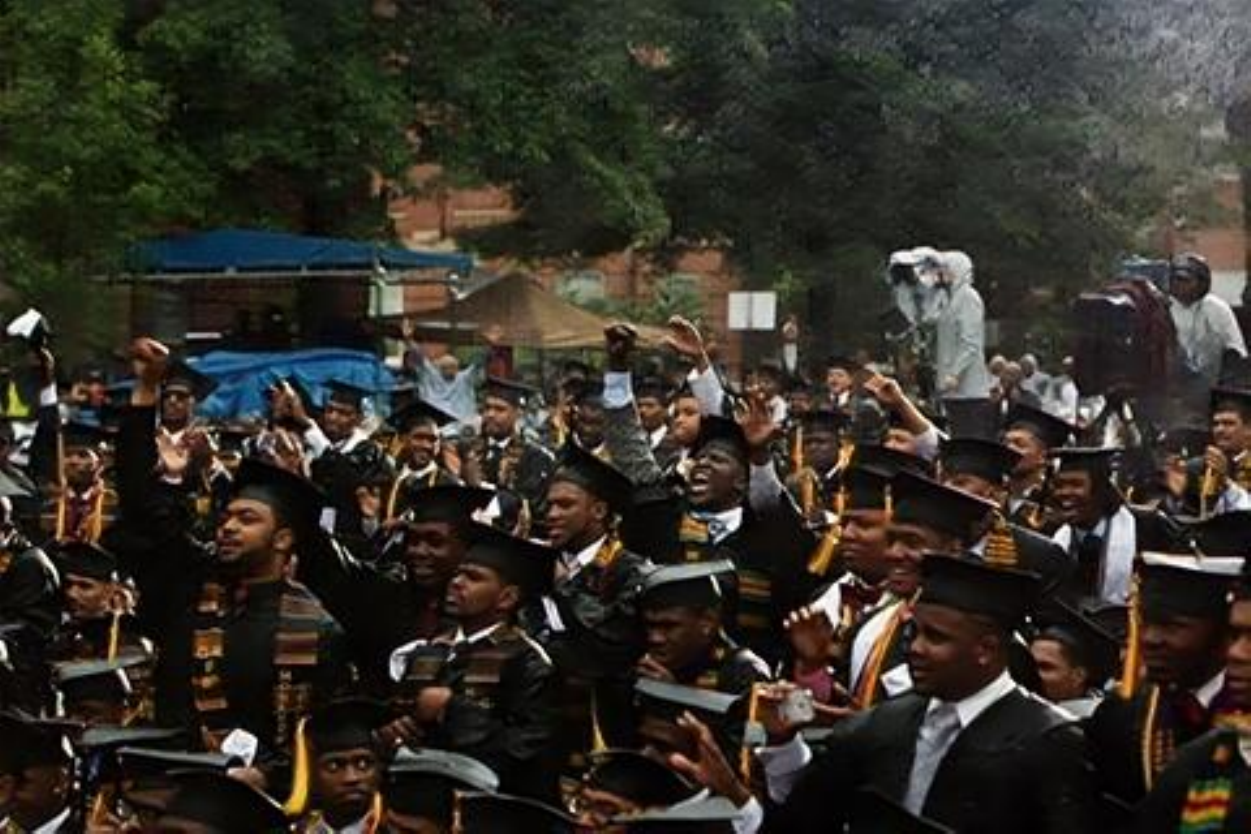}
}
\caption{Image rain removal results on a real-world heavy rainy image. From (a) to (f): (a) the real rainy image, (b-c) the supervised learning results, (d-e) the semi-supervised learning results, (f) the rain removal + haze removal results, (g) DGNL-Net, and (h) our Semi-MoreGAN, respectively. Semi-MoreGAN generates both rain-free and perceptually more pleasing results with image details preserved.}
\label{fig:intro_real}
\end{figure*}
In real-world rain scenes, we observe that rain is a mixture of rain streaks and rainy haze, which commonly appears in various forms under different shooting conditions: On the background scene, the rain occlusions closer to the camera are resulted from the rain streaks while those far away are caused by the rainy haze that follows with the rain streaks. However, current learning-based rain removal methods lack full consideration of the mixture of rain, and only focus on removing single types of rain artifacts, i.e., rain streaks or rainy haze. The entanglement of rain streaks and rainy haze inevitably degrades their rain removal abilities to deal with heavy rainy images (see Fig. \ref{fig:intro_real} (b)-(e)). Additionally, most of them are trained with paired synthetic data to alleviate the difficulty of collecting real-world rainy/clean image pairs. These synthetic rainy images cannot include a sufficiently comprehensive range of rain streaks patterns like real-world rainy images, such as direction and density. The distributional-shift between synthetic and real-world rainy images leads to the sub-optimal performance on the real-world rainy images (see Fig. \ref{fig:intro_real} (f)-(g)). Differently, we propose a semi-supervised learning paradigm 
that effectively removes rain streaks and rainy haze together with the guidance of the predicted accurate depth map, and leverages real-world rainy/clean image pairs to improve the rain removal ability when facing the real-world rain (see Fig. \ref{fig:intro_real} (h)).

In this paper, we propose a novel \textbf{SEMI}-supervised \textbf{M}ixture \textbf{O}f rain \textbf{RE}moval \textbf{G}enerative \textbf{A}dversarial \textbf{N}etwork (Semi-MoreGAN) for mixture of rain removal. 
Extensive experiments demonstrate the superiority of Semi-MoreGAN in removing the intricately entangled degradations under heavy rainy conditions. Also, our approach qualitatively and quantitatively outperforms existing supervised and semi-supervised rain removal algorithms on both synthetic and real-world images.
The main contribution of this work can be concluded as follows:
\begin{itemize}
\item[$\bullet$]
Beyond existing supervised rain removal efforts, 
the proposed Semi-MoreGAN consists of a supervised branch and an unsupervised branch, to leverage both synthetic datasets and real-world images for the network's training, thus can well address the unpredictable real-world rainy scenes.
\item[$\bullet$]
We propose a novel attentional depth prediction network (ADPN) with the help of self-attention mechanism for accurate depth map prediction, which guides Semi-MoreGAN to remove nearby rain streaks and far rainy haze together.
\item[$\bullet$]
We propose a contextual feature prediction network (CFPN) with well-designed contextual feature aggregation blocks, which can enlarge the receptive fields to enrich the output features.
%
\item[$\bullet$]
We propose a pyramid depth-guided non-local network (PDNL) 
to learn non-local features from contextual features with the guidance of the depth map, which enables to produce the promising results in removing mixture of rain. 
%

\end{itemize}

\section{Related Work}
Current studies can be categorized into two groups according to the rain forms: \textit{rain streaks removal}, and simultaneous \textit{rain streaks} and \textit{rainy haze removal}. The majority emphasizes handling rain streaks of disparate directions and densities, while Li et al.~\cite{li2019heavy} for the first time investigate the heavy rain removal.

\subsection{Rain Streaks Removal}
Rain dynamically falling in the air is imaged as streaks, which are in similar directions and sparsely distributed in the whole image~\cite{garg2007vision}. Many studies exploit these physical properties to build various image priors in the optimization framework. In~\cite{p_kang2011automatic}, it is assumed that rain streaks are sparse and of high frequency in the image. Thus they exploit histogram of oriented gradients (HOGs) features of rain streaks to cluster into rain and non-rain dictionary. In~\cite{p_du2018single}, the low-rank representations are incorporated into rain streaks removal. 
Luo et al.~\cite{luo2015removing} propose a novel sparse coding scheme considering the discrimination of background and rain streaks. 
Li et al.~\cite{li2016rain} adopt the Gaussian Mixture Model (GMM) to accommodate multiple orientations and scales of rain streaks. 


The introduction of neural networks significantly boosts the rain removal performance significantly. Fu et al.~\cite{d_detail_layer} first make an attempt to remove rain streaks via a deep detail network. Thereafter, many efforts have been made to either introduce advanced network modules and structures, or integrate problem-related knowledge into network design. Network modules, such as dense block~\cite{fan2018residual}, recursive block~\cite{ren2019progressive} and dilated convolution~\cite{li2018recurrent, yang2017deep,liang2019rain,ren2020scga,ren2020not,deng2020detail}, and structures, such as RNN~\cite{li2018recurrent,ren2019progressive}, GAN~\cite{li2019heavy,zhang2019image} and multi-stream networks~\cite{yang2017deep,li2018recurrent}, are validated to be effective in rain streaks removal. Auxiliary information, including rain density~\cite{zhang2018density}, streak position~\cite{yang2017deep} and gradient information~\cite{d_wang2019gradient}, are leveraged to improve the robustness and performance of rain removal networks.

To improve the generalization of real-world image rain removal, Wei et al.~\cite{wei2019semi} first propose a semi-supervised rain removal framework where they adopt a likelihood term imposed on  Gaussian Mixture Model (GMM) and minimize the Kullback-Leibler (KL) divergence between synthetic and real rain. Yasarla et al. \cite{yasarla2020syn2real} employ Gaussian Process (GP) to generate pseudo-labels to model the real rain. Recently, Ye et al. \cite{ye2021closing} propose a bidirectional disentangled translation network by decomposing the rainy image into a clean background and rain layer. However, these methods only focus on removing rain streaks and ignore the physical properties of rainy haze, thus achieving sub-optimal performance on heavy rainy images.

\subsection{Rain Streaks \& Rainy Haze Removal}
Studies on rainy haze removal~\cite{he2010single,zhang2018densely,qin2020ffa,qu2019enhanced} were independent of rain streaks removal until Li et al.~\cite{li2019heavy} observe that rain in the distance of a scene actually generates haze-like effects, and thus they design a depth-guided GAN to handle both the rain and haze. Later, several studies attempt to remove both rain streaks and rainy haze from the images \cite{hu2021single,wang2020rethinking,hu2019depth}. For example, Hu et al.~\cite{hu2021single} present a depth-guided network, which utilizes the depth feature to produce a spatial attention map that guides the process of rain removal. A similar scheme is utilized by Wang et al.~\cite{wang2020rethinking} but with a different mixture rain formulation. However, these methods are still trained with synthetic datasets, and cannot effectively integrate the representations of depth features and image features, which degrade their rain removal ability in real-world scenes.

\begin{figure}[!t]
	\centering
	\includegraphics[width=1.0\linewidth]{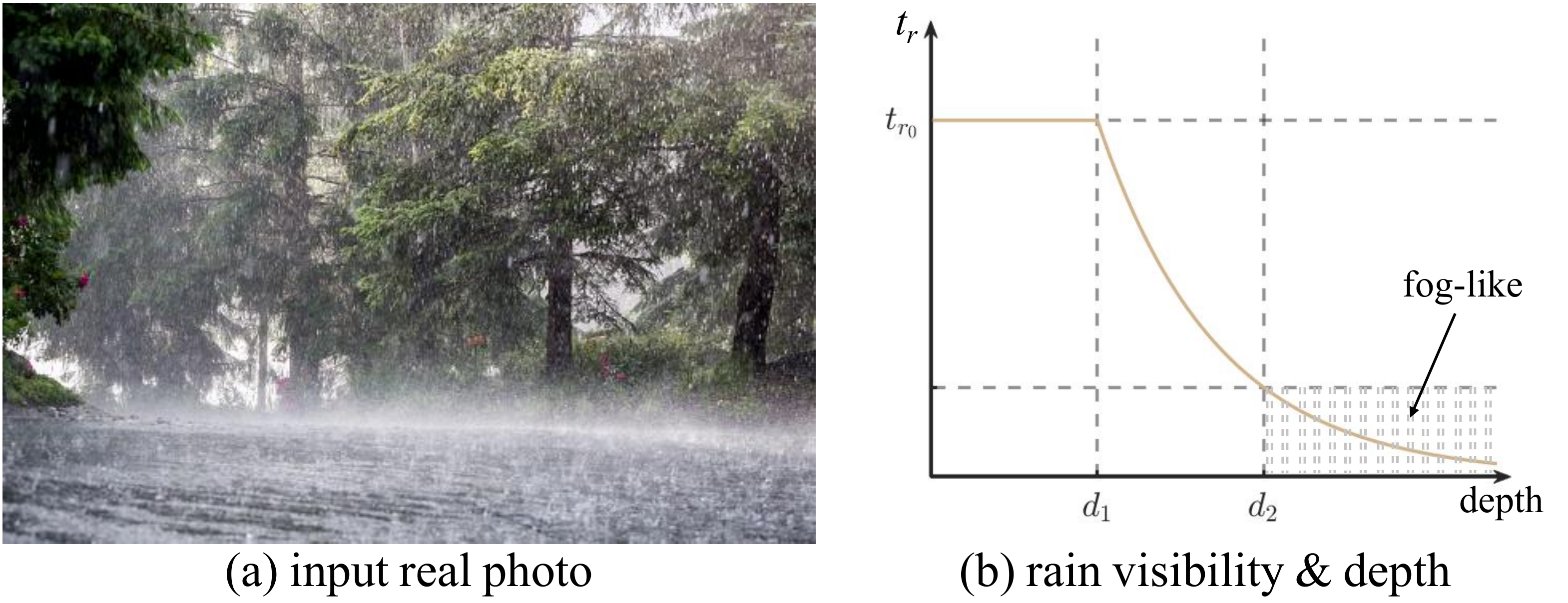}
    \caption{Real-world rain is a mixture of rain streaks and rainy haze, and the scene depth determines the transformation from rain streaks to rainy haze. (a) Nearby is rain streaks; far away is rainy haze.  (b) Plot of rain streaks intensity $(t_{r})$ against the scene depth $(d)$ based on the model in \cite{garg2007vision}.}
	\label{fig:depthrainmodel}
\end{figure}
\begin{figure*}[!ht]
	\centering
	\includegraphics[width=0.9\linewidth]{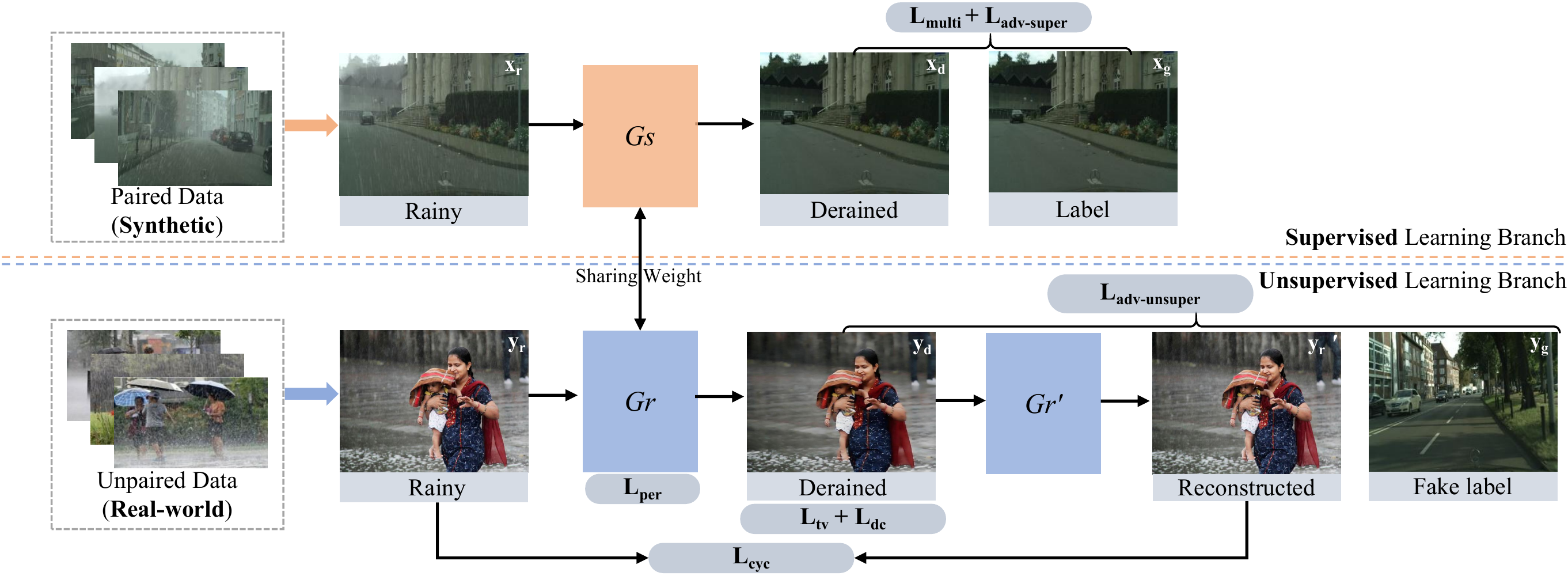}
    \caption{Overview of \textbf{SEMI}-supervised \textbf{M}ixture \textbf{O}f rain \textbf{RE}moval \textbf{G}enerative \textbf{A}dversarial \textbf{N}etwork (Semi-MoreGAN). The training process consists of a supervised learning branch and an unsupervised learning branch. In particular, $x_{r}$ and $y_{r}$ are the rainy image from the synthetic dataset and real-world dataset, respectively. $x_{d}$ and $y_{d}$ are derained images, $y_{r}'$ is reconstructed rainy image, $x_g$ is the corresponding clean image of $x_r$, and $y_{g}$ is the fake label, which is randomly chosen from the synthetic dataset as the ground truth of unsupervised learning branch.}
	\label{fig:overview}
\end{figure*}
\section{Motivation}
Real-world rain is a mixture of rain streaks and rainy haze.
It is observed from Fig. \ref{fig:depthrainmodel}(a) that, rain streaks become strongly visible nearby, and parts of them will accumulate and change to rainy haze far away. In the captured image of real-world rain, rain streaks and rainy haze will be entangled with each other. It is more reasonable to consider designing a mixture of rain removal model, instead of the simplified version, i.e., only rain streaks \cite{wang2019spatial,li2019semi,zamir2021multi,zhang2019image}, or rainy haze. From Fig. \ref{fig:depthrainmodel} (b) we know, formulating an imaging model is possible for simultaneous rain streaks and rainy haze removal, if the scene depth can be well estimated. 

Furthermore, to bridge the domain gap between synthetic and real-world rainy images, an effective image rain removal paradigm should leverage both synthetic and real-world data to enhance the model's generalization ability. Thus, it is desirable to develop a semi-supervised network that first trains itself on synthetic data and then transfers the pre-trained model to real-world data.

\section{Semi-MoreGAN}
The rainy image is the combination of rain streaks, rainy haze, image background, and the light effect; based on the fact we formulate a rainy imaging model. To solve the model, we propose a \textbf{SEMI}-supervised \textbf{M}ixture \textbf{O}f rain \textbf{RE}moval \textbf{G}enerative \textbf{A}dversarial \textbf{N}etwork (termed Semi-More GAN). 
Semi-MoreGAN leverages both synthesized and real-world rainy images to train itself in a semi-supervised manner. Therefore, Semi-MoreGAN can bridge the domain shift gap and generalize well in real-world scenarios. The architecture of Semi-MoreGAN is illustrated in Fig. \ref{fig:overview}.

In the following, we first formulate the rain imaging model by considering both rain streaks and rainy haze. After that, we elaborate on our generator and discriminator, to demonstrate how we dissolve the intractable problem of mixture of rain removal. Finally, we present the loss functions that make the results of real-world rain removal  clearer and more realistic.

\subsection{Rain Image Formulation}
Rain degradation is a very complex process, and there are several studies on it in the field of computer graphics, that is, how to 
synthetically render rain degradation and generate rainy images \cite{rousseau2006realistic,wang2006real,tatarchuk2006artist,garg2006photorealistic,wang2020model}. Motivated by these approaches, we formulate a rain imaging model for better rain removal.
According to Garg and Nayar~\cite{garg2006photorealistic}, the visual intensity of the rain streaks and the rainy haze layers depends on the scene depth from the camera to the underlying scene objects behind the rain. Thus, we formulate the rain streaks layer $S(x)$ as:
\begin{equation}\label{eq:depth}
S(x)=S(x)_\textbf{pattern} \ast \textbf{e}^{-\alpha\textbf{max}(d_1,d(x))}
\end{equation}
where $S(x)_\textbf{pattern}\in[0,1]$ is an intensity image of uniformly-distributed rain streaks in the image space, $\ast$ denotes the pixel-wise multiplication, $\alpha$ is an attenuation coefficient that controls the rain streaks intensity, $d(x)$ means the scene depth~\cite{garg2006photorealistic}.

Meanwhile, according to the standard optical model~\cite{koschmieder1924theorie} that simulates the image degradation process, the visual intensity of rainy haze increases exponentially with the scene depth. Hence, we model the rainy haze layer $A(x)$ as:
\begin{equation}\label{eq:fogdepth}
A(x)=1 - \textbf{e}^{-\beta d(x)}
\end{equation}
where $\beta$ is an attenuation coefficient that controls the thickness of rainy haze. A larger $\beta$ means a thicker rainy haze and vice versa.

An image degraded by rain is composed of a background image and a mixture of two layers, i.e., the rain streaks layer and the rainy haze layer. Similar to \cite{hu2019depth,li2019heavy,wang2020rethinking,hu2021single}, we devise the captured rainy image $I(x)$ at a pixel $x$ as:
\begin{equation}\label{eq:mixture_rain}
\begin{aligned}
I(x)= B(x)(1-S(x)-A(x))+S(x)+\hat{\alpha} A(x)
\end{aligned}  
\end{equation}
where $B(x)$ denotes the clean background image with the clear scene radiance, $S(x)\in[0,1]$ and $A(x)\in[0,1]$ are the rain streaks layer and the rainy haze layer, respectively. $\hat{\alpha}$ denotes the atmospheric light, which is assumed to be a global constant~\cite{sakaridis2018semantic}.


\subsection{Network Architecture}
\begin{figure*}[!ht]
	\centering
	\includegraphics[width=1\linewidth]{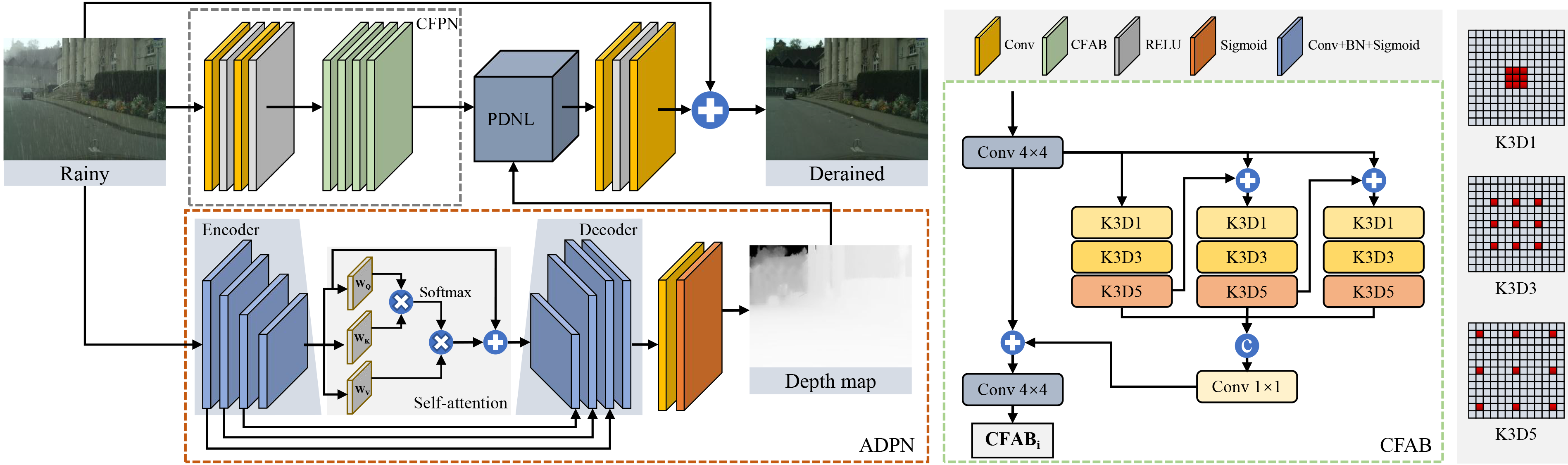}
	\caption{Generator. It consists of (i) an attentional depth prediction network (ADPN) to predict more accurate depth maps based on the self-attention mechanism; (ii) a contextual feature prediction network (CFPN) to extract contextual features from the rainy image; (iii) a pyramid depth-guided non-local network (PDNL) to generate non-local features in a depth-guided manner; (iv) several convolution layers with a residual connection to obtain the final derained image.}
	\label{fig:generator}
\end{figure*}

\begin{figure}[!t]
	\centering
	\includegraphics[width=1\linewidth]{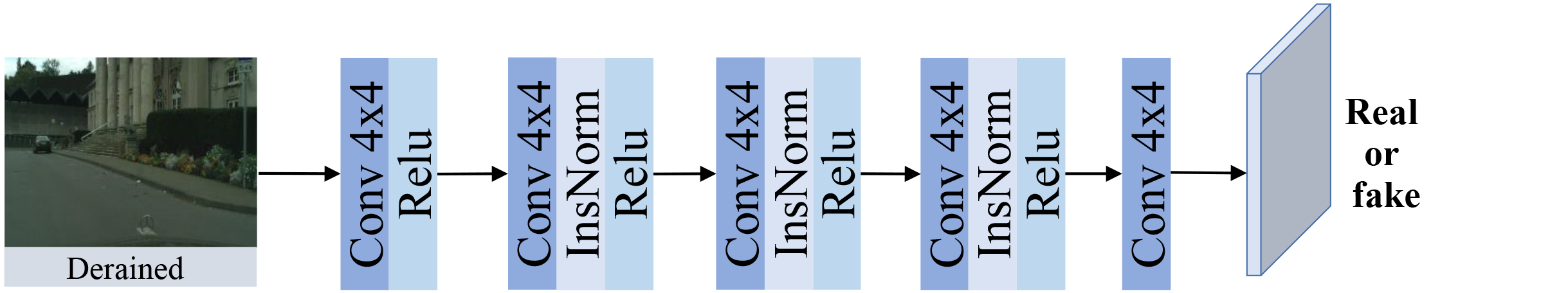}
	\caption{Discriminator. It guides the generator to produce higher-quality derained images.}
	\label{fig:discriminator}
\end{figure}

The architecture of Semi-MoreGAN is illustrated in Fig. \ref{fig:overview}. Different from existing rain removal approaches \cite{hu2021single,li2019heavy,wang2020rethinking,hu2019depth} trained only on synthetic data, our Semi-MoreGAN is a semi-supervised paradigm, which can leverage unpaired real-world rainy and clean images to bridge the gap between synthetic and real-world rain.
It consists of supervised and unsupervised branches. 
Semi-MoreGAN has three generators ($Gs, Gr, Gr'$) and two discriminators ($Ds, Dr$), where $Gs$ and $Gr$ share the weights during training. In the supervised training phase, we input the synthetic rainy image $x_{r}$ into $Gs$ to obtain the derained image $x_{d}$, and utilize the supervised loss function consisting of multi-task loss and adversarial loss to guide training. In the unsupervised training phase, we feed the real-world rainy image $y_{r}$ into $Gr$ to obtain the derained image $y_{d}$, then $y_{d}$ is fed into $Gr'$ to obtain the reconstructed rainy image $y_{r'}$. Since the real-world image $y_{r}$ has no corresponding clean image, we adopt the unsupervised loss function consisting of dark channel loss, TV loss, cycle-consistency loss, perceptual loss, and adversarial loss, to produce real-world derained images without the supervision of paired data. Since $Gs$ and $Ds$ have the same architecture as $Gr,Gr'$ and $Dr$, respectively, we only present the architectures of $Gs$ and $Ds$ (see Fig. \ref{fig:generator} and Fig. \ref{fig:discriminator}).

\textbf{Generator}. The detailed architecture of generator is shown in Fig. \ref{fig:generator}. Given a rainy image, our goal is to estimate the corresponding rain-free image as output in an end-to-end manner. The generator consists of three sub-networks: (i) in light that depth information is crucial for mixture of rain removal, we develop a novel attentional depth prediction network (ADPN) to predict more robust and sharper depth maps; (ii) a contextual feature prediction network (CFPN) is proposed to extract global contextual features; (iii) since such contextual features are still local, we propose a pyramid depth-guided non-local network (PDNL) with multi-scale sampling for the depth-guided non-local features extraction. Finally, 
we employ two convolution layers followed by a non-linear ReLU layer with a residual connection to obtain the final derained image.

\begin{figure}[t]
\begin{center}
  \includegraphics[width=1\linewidth]{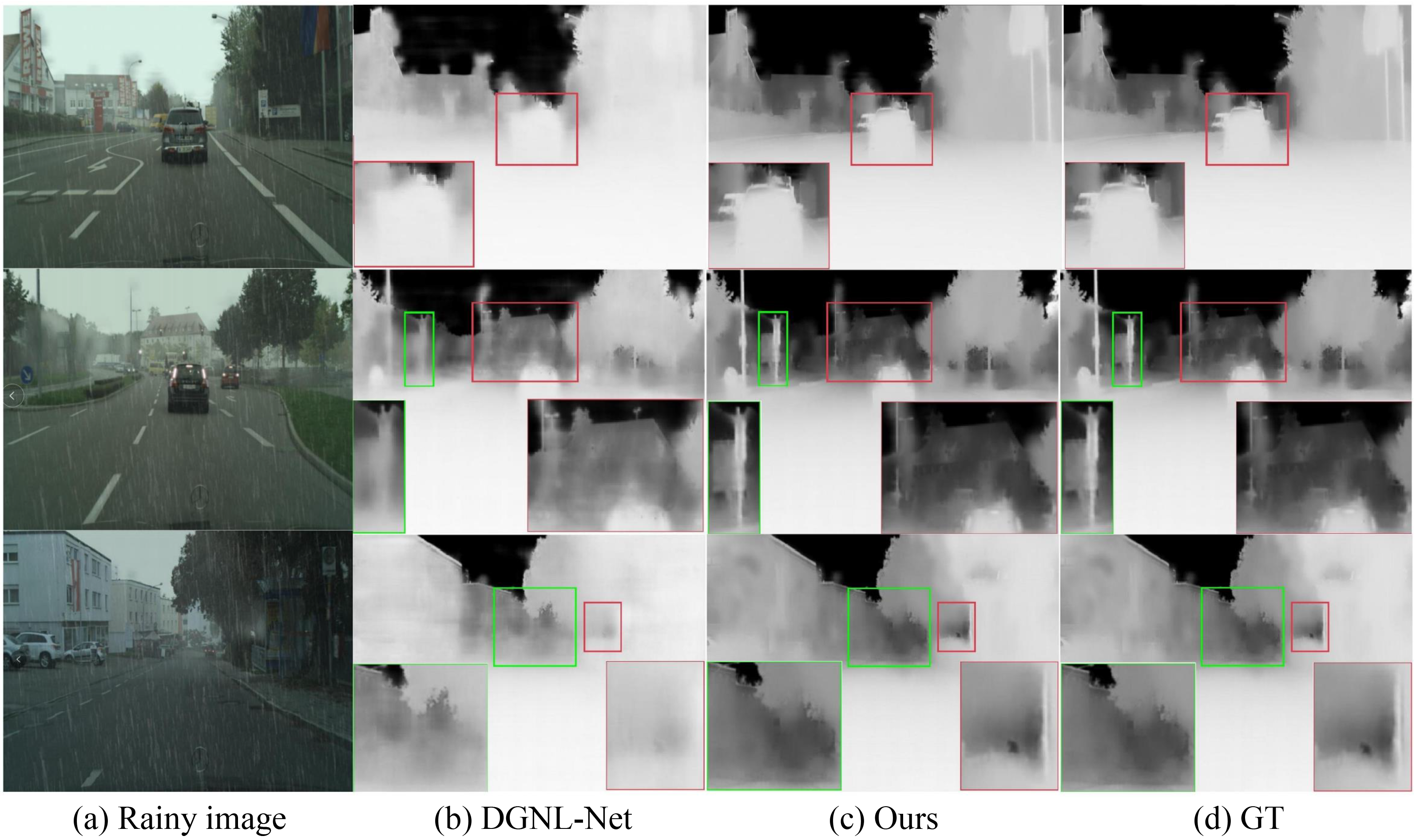}
	\caption{Visualizing the predicted depth maps by DGNL-Net~\cite{hu2021single} and our Semi-MoreGAN.}
	\label{depth}
	\end{center}
\end{figure}
\textbf{Attentional depth prediction network}.
In heavy rain scenes, real-world rain is a mixture of rain streaks and rainy haze. As shown in Fig. \ref{fig:depthrainmodel}(b), the transformation from rain streaks to rainy haze depends on the scene depth. Thus, it is reasonable to provide additional depth estimation to guide the rain removal model to remove both nearby rain streaks and far rainy haze. However, coarse and inaccurate depth estimation will introduce interfering information to disturb the process of mixture of rain removal, thus leading to sub-optimal performance.

To obtain a more robust and sharper depth map, we present the attentional depth prediction network (ADPN), which employs an auto-encoder structure with skip connections between
the encoder and the decoder. Concretely, the encoder takes a rainy image as input, and the decoder outputs its depth map. They consist of eight convolution layers followed by batchnorm and sigmoid layers together (see Fig. \ref{fig:generator}). It is noteworthy that the final depth prediction is obtained after a convolution layer and a sigmoid layer.
Further, we explore adopting a self-attention block \cite{vaswani2017attention} between encoder and decoder, which learns to contextually represent groups of features with similar semantic similarities. We argue that the self-attention mechanism can improve depth prediction, since the correct context for the prediction of a pixel depth map may be at a non-contiguous location that the normal convolution operations cannot reach. Concretely, the self-attention block first takes $F_{d}$ as input to compute the query $Q$, the key $K$ and the value $V$:
\begin{equation}\label{eq:freq_com}
\begin{aligned}
Q=W_{Q}F_{d},\\
K=W_{K}F_{d},\\
V=W_{V}F_{d}\\
\end{aligned}
\end{equation}
where $W_{Q}$, $W_{K}$ and $W_{V}$ are learned parameters, and the self-attention and original features jointly contribute to the output by
\begin{equation}\label{eq:freq_coms}
S_{F}=Softmax(F_{d}^{T}W_{Q}^{T}W_{K}F_{d})V+F_{d}
\end{equation}
where $S_{F}$ represents the output of the self-attention block, which is fed into the decoder to recover the image resolution.
%
Although DGNL-Net \cite{hu2021single} also designs a depth prediction network for rain streaks and rainy haze removal, its CNN-based depth prediction network cannot guarantee the accuracy of the generated depth map, and local convolution cannot capture the context information of the whole image. Different from it, our ADPN can provide a more accurate depth prediction by exploiting the self-attention mechanism, as shown in Fig. \ref{depth}. 
%

\textbf{Contextual feature prediction network.}
Since a large receptive field plays an important role in obtaining more context information, which is helpful for automatically identifying rain patterns and removing the mixture of rain. Thus we design a contextual feature prediction network (CFPN) with four well-designed contextual feature aggregation blocks (CFABs) as shown in Fig. \ref{fig:generator}. CFAB employs several multi-scale dilated blocks to enlarge the receptive fields successively. Concretely, we first adopt a 4 × 4 convolution layer to obtain the input feature $F_{c}$, which is fed into three branches to capture different characterizations. Each branch is equipped with three dilated convolution layers whose dilation scales are 1, 3, and 5 for acquiring much contextual information, and the output of each branch is added to the input of the next branch, which can be expressed as:
\begin{equation}\label{eq:CFABI}
\begin{aligned}
Cout_{k}^{i}=\left\{
\begin{array}{rcl}
Conv_{d}(F_{c}^{i})      &      {k = 1}\\
Conv_{d}(F_{c}^{i} + Cout_{k-1}^{i})      &      {k=2,3}
\end{array} \right. 
\end{aligned}
\end{equation}
where $F_{c}^{i}$ denotes the $ith$ feature map produced by the CFAB, $Cout_{k}^{i}$ denotes the output of the $kth$ branch, $Conv_{d}$ presents the stacked dilated convolutional layers. Then, we concatenate the outputs of three branches followed by a 1 × 1 convolution layer to reduce the channel, and add it to the input feature $F_{c}^{i}$. Finally, we adopt a 4 × 4 convolution layer to obtain the final contextual feature map $CFAB_{i},i\in \{ 1,2,3,4\}$ which is computed as:
\begin{equation}\label{eq:RD}
\begin{aligned}
CFAB_{i} = Conv(F_{c}^{i}+Conv(Cat_{k=1}^{3}(Cout_{k}^{i})))
\end{aligned}
\end{equation}
where $Cat_{k=1}^{3}$ denotes the concatenation of all three branches. The overall structure of CFAB is shown in Fig. \ref{fig:generator}.

\begin{figure}[!t]
	\centering
	\includegraphics[width=1\linewidth]{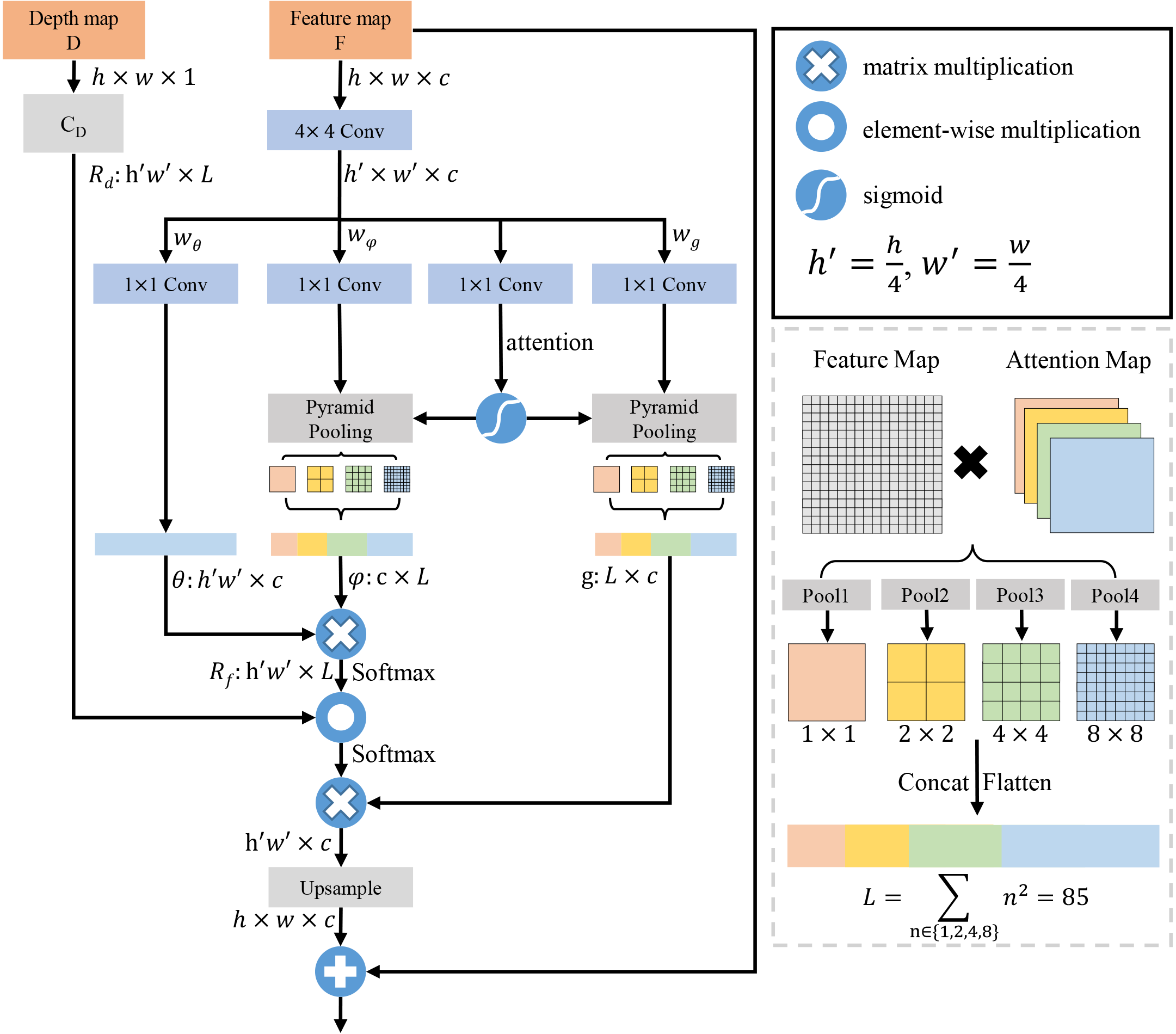}
	\caption{Pyramid Depth-guided Non-local Network (PDNL).}
	\label{fig:PDNL}
\end{figure}

\textbf{Pyramid depth-guided non-local network.}
Although the CFPN enables to aggregate context information at multiple scales, the extracted contextual features are still local \cite{wang2018non}. To this end, we develop a pyramid depth-guided non-local network (PDNL) as shown in Fig. \ref{fig:PDNL}. To generate depth-guided non-local features, the proposed PDNL captures the long dependency between depth features and image features and provides a more effective fusion manner.

The inputs include depth map $D\in R^{h\times w \times 1}$ and contextual feature map $F\in R^{h\times w \times c}$. We first reshape $D$ to a $1\times hw$ vector and adopt the operation $C_{D}$ \cite{hu2021single} to compute the depth distance in logarithmic space between each pair of pixels on the depth map, and obtain the depth relation map $R_{d} \in R^{h'w'\times L}$ after downsampling, which can be expressed as:
\begin{equation}\label{eq:RD}
\begin{aligned}
R_{d}[i][j]&= Downsample(Softmax(min(\frac{D_{i}}{D_{i}+\epsilon},\frac{D_{j}}{D_{j}+\epsilon})))
\end{aligned}
\end{equation}
where $[i,j]$ represents the position $(i,j)$, $\epsilon$ is a small value set as $1\times e^{-6}$ to avoid division by zero and $h'= \frac{h}{4}$, $w'= \frac{w}{4}$. 
Meanwhile, we adopt a 4 × 4 convolution layer to reshape the contextual feature map from $F\in R^{h\times w \times c}$ to $F\in R^{h'\times w' \times c}$. Then, we measure the feature distance between each pair of pixels to obtain the feature relation map $R_{f}$. 
In particular, we develop three operations $W_{\theta}$, $W_{\varphi}$ and $W_{g}$ (which both include 1$\times$1 convolution layers and pyramid pooling layers) to obtain feature map $\theta\in R^{h'w' \times c}$, $\varphi\in R^{c\times L}$ and $g\in R^{L \times c}$. Then the feature relation map $R_{f}$ can be computed by:
\begin{equation}\label{eq:FD}
\begin{aligned}
R_{f}&= Softmax(F^{T}W_{\theta}^{T}W_{\varphi}F)
\end{aligned}
\end{equation}
Once $R_{d}$ and $R_{f}$ are obtained, we use an element-wise multiplication and softmax layer on them, then the fused result is multiplied with the feature map $g$, and added to the input feature map $F$ after upsampling,
to obtain the output of PDNL (denoted as $F_{o}$), which is expressed as :
\begin{equation}\label{eq:FD2}
\begin{aligned}
F_{o} = Upsample(g\otimes Softmax(R_{d}\odot R_{f}))+F
\end{aligned}
\end{equation}
where $\odot$ and $\otimes$ represent element-wise multiplication and matrix multiplication, respectively. 

\textbf{More details about both the similarity and difference between our PDNL and DGNL \cite{hu2021single}.} The similarity between the proposed PDNL and depth-guided non-local module (DGNL) \cite{hu2021single} is the use of the standard non-local module, which is first proposed by \cite{wang2018non} and utilized to generate non-local features. While the proposed PDNL provides a more effective fusion manner, the differences lie in several aspects. 

Although DGNL is promising in establishing long-range dependencies, its computational complexity is $O(N^{2}C)$, which is very computationally expensive and inefficient compared to normal convolutions. Where $N=h\times w$, and $h$, $w$, and $C$ represent the spatial height, width, and channel number of the feature map, respectively. 
To reduce the computational cost and enhance the global and multi-scale representations, we embed pyramid pooling layers \cite{zhao2017pyramid} into the non-local blocks, where the number of features after sampling reduces to $L$, thus the complexity of PDNL is $O(NLC)$. This sampling process is depicted in Fig. \ref{fig:PDNL}. For example, we set the pooling size $n \in \{1,2,4,8\}$ on the feature map $64 \times 128$, then four pooling results are flattened and concatenated to serve as the input to the next layer. Obviously, the total number of features $L = 85$ after downsampling is much smaller than $N = 8192$. 

However, the pyramid pooling operations may lead to the image resolution problem: the low-resolution features would be replaced with high-resolution features during the downsampling process. To this end, we design an additional attention module, which contains a $1 \times 1$ convolution layer to capture spatial attention maps as weights. And this module adaptively adjusts the weights to focus more attention on useful information. Such a weighted pyramid pooling operation provides a more effective way to gather the key information of the origin feature map for fusion while reducing computational complexity.

\textbf{Discriminator}. For adversarial training, the discriminator accepts an output image produced by the generator and focuses on more local areas that are likely to be fake by the given images $x_{d}$ or $y_{d}$. As demonstrated in Fig. \ref{fig:discriminator}, the discriminator consists of five 4×4 convolution layers, three instance normalization layers, and four non-linear ReLU layers, and outputs a $64 \times 64$ patch, which is utilized to check whether the input image is a real image or a fake image generated by $G_s$ or $G_r$. 
\begin{figure*}[!ht]
	\centering
	\includegraphics[width=1\linewidth]{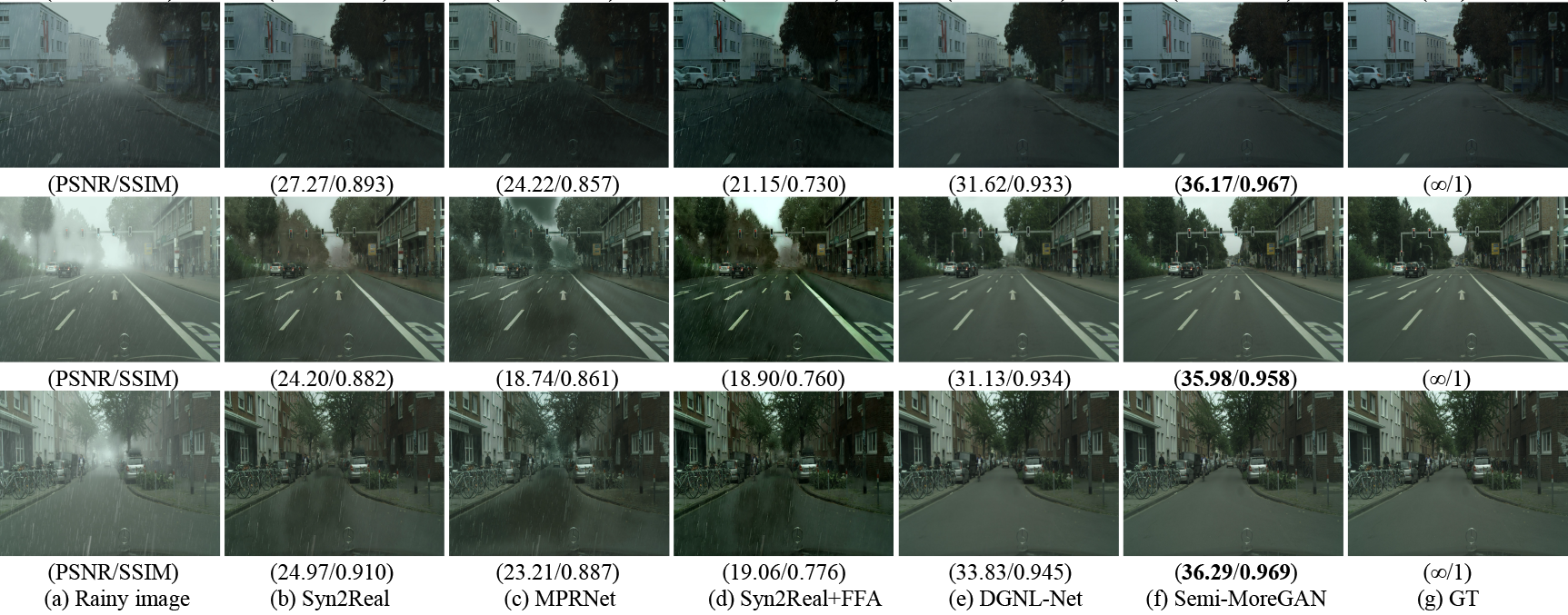}
	\caption{Qualitative evaluation on RainCityscapes dataset. From (a) to (g): (a) Rainy image, and the rain removal results of (b) Syn2Real~\cite{yasarla2020syn2real}, (c) MPRNet \cite{zamir2021multi}, (d) Syn2Real~\cite{yasarla2020syn2real}+FFA~\cite{qin2020ffa}, (e) DGNL-Net~\cite{hu2021single}, (f) our Semi-MoreGAN, and (g) Ground Truth (GT). In mixture of rain, nearly all rain streaks and rainy haze are removed by our method, despite the diversity of their colors, shapes and transparency, while others are commonly not.}
 \label{fig:f1}
\end{figure*}

\subsection{Loss Function}
Different from existing rain removal approaches, Semi-MoreGAN is trained in a semi-supervised manner. We consider both supervised and unsupervised loss functions to improve the quality of final rain removal images. During training, the multi-task loss, adversarial loss, cycle-consistency loss, dark channel (DC) loss, perceptual loss and the total variation (TV) loss are employed, which can be formulated as:
\begin{equation}
\begin{split}
L_{Total}=&{\lambda_{1}L_{multi}+\lambda_{2}L_{adv-super}(Gs)+\lambda_{3}L_{cyc}}\\
&{+\lambda_{4}L_{adv-unsuper}(Gr)+\lambda_{5}L_{dc}+\lambda_{6}L_{tv}+\lambda_{7}L_{per}}
\end{split}
\end{equation}

\textbf{Multi-task loss}. We adopt a multi-task loss function for jointly optimizing the depth estimation and rain removal. Especially, given a synthetic rainy image $x_{r}$, we adopt the $L_1$ loss to ensure the rain removal image $Gs(x_{r})$ and depth map $d$ generated by ADPN, are close to the ground truth $x_g$ and $d_{g}$ by:
\begin{equation}
L_{multi} = \parallel Gs(x_{r})-x_g\parallel_1+ \parallel d-d_g\parallel_1
\end{equation}

\textbf{Adversarial loss}. To make the final generated images ($x_{d}$ and $y_{d}$) much clearer and more realistic, we build two discriminators to distinguish whether an input image is produced by the generator or a clean image. Here we adopt Least Squares GAN (LSGAN) \cite{mao2017least} to calculate the adversarial loss as:
\begin{equation}\label{eq:advg-su}
\begin{aligned}
L_{adv-super}(Gs)=E_{Gs(x_r) \sim P_{fake}}[(Ds(Gs(x_r))-1)^2]
\end{aligned}
\end{equation}
\begin{equation}\label{eq:advd-su}
\begin{aligned}
L_{adv-super}(Ds)=E_{x_g \sim P_{real}}[(Ds(x_{g})-1)^2]\\
+E_{Gs(x_r) \sim P_{fake}}[(Ds(Gs(x_r)))^2]
\end{aligned}
\end{equation}
\begin{equation}\label{eq:advg-un}
\begin{aligned}
L_{adv-unsuper}(Gr)=E_{Gr(y_r) \sim P_{fake}}[(Dr(Gr(y_r))-1)^2]
\end{aligned}
\end{equation}
\begin{equation}\label{eq:advd-un}
\begin{aligned}
L_{adv-unsuper}(Dr)=E_{y_g \sim P_{real}}[(Dr(y_{g})-1)^2]\\
+E_{Gr(y_r) \sim P_{fake}}[(Dr(Gr(y_r)))^2]
\end{aligned}
\end{equation}
where $x_{r}$ and $x_{g}$ are rainy images and corresponding clean images from the synthetic dataset, respectively, $y_{r}$ denotes real-world rainy images. It is noteworthy that since there is no corresponding clean image for most real-world rainy images and hence, we use the fake label $y_{g}$ as the corresponding clean image of $y_{r}$, which is randomly chosen from the synthetic dataset, to constrain the adversarial loss in the unsupervised learning branch.

\textbf{Cycle-consistency loss}. For the unsupervised learning branch, we employ cycle-consistency loss \cite{zhu2017unpaired} to make the mixture of rain removal images possess similar feature distributions with the real-world clean images. The $L_1$ loss is adopted to compute the cycle-consistency loss, making the unsupervised training process possible:
\begin{equation}
\begin{split}
L_{cyc}={\parallel}\text{(}y_{r}'-y_{r}\text{)}\mathop{ {\parallel} }\nolimits_{{1}}
\end{split}
\end{equation}
where $y_{r}'$ refers to the reconstructed rainy image from generator $Gr'$. This loss function is designed to encourage the reconstructed image to match the input image closely, i.e., $y_{r}' {\approx} y_r$.

\textbf{Dark channel loss}. Since dark channel prior has been proved very efficient to remove the haze from hazy images. Inspired by~\cite{li2019semi}, we adopt the dark channel loss to deal with the rainy haze in the real-world images as:
\begin{equation}
L_{dc} = \parallel \textbf{D}_{Gr(y_r)}{\parallel}_1
\end{equation}
where $\textbf{D}_{Gr(y_r)}$ denotes the dark channel vector form of the mixture of rain removal image $Gr(y_r)$. 

\textbf{Total variation loss}. To preserve both structures and details from input images, we introduce the total variation (TV) loss \cite{aly2005image} to enforce spatial smoothness of rain removal images as:
\begin{equation}
\begin{split}
L_{tv}= {\parallel} \mathop{ \nabla }\nolimits_{{x}}\mathop{{G}}\nolimits_{{r}}\text{(}y_r\text{)}+\mathop{ \nabla }\nolimits_{{y}}\mathop{{G}}\nolimits_{{r}}\text{(}y_r\text{)}\mathop{ {\parallel} }\nolimits_{{1}}
\end{split}
\end{equation}
where $\nabla_{x}$ and $\nabla_{y}$ represent the horizontal and vertical differential operation matrices, respectively. This loss function can remove noise from the produced images, thus making them much clearer.

\textbf{Perceptual loss}. To preserve both large-scale structures and small-scale details from rainy input, we adopt a perceptual loss \cite{johnson2016perceptual} to calculate perceptual similarity as:
\begin{equation}
\begin{split}
L_{per}= \parallel VGG(y_{d})-VGG(y_r)\parallel_2^{2}
\end{split}
\end{equation}
where $VGG(.)$ denotes the feature maps extracted from the $2^{nd}$ and $5^{th}$ pooling layers within the VGG-16 network pre-trained on ImageNet. This perceptual loss can make the derained image have similar structure and texture features to the input image.

\section{Experiments}
\subsection{Experimental Settings}
We implement Semi-MoreGAN using Pytorch 1.6 on a system with 11th Gen Intel(R) Core(TM) i7-11700F CPU and Nvidia GeForce RTX 3090 GPU. For optimizing Semi-MoreGAN, we employ the Adam optimizer with the first momentum value of 0.9, the second momentum value of 0.999, and a weight decay of zero. Besides, the initial learning rates for generators and discriminators are set to $5e^{-4}$ and $1e^{-5}$, respectively. In the experiment, we set $\lambda_{1}$, $\lambda_{2}$, $\lambda_{3}$, $\lambda_{4}$, $\lambda_{5}$, $\lambda_{6}$ and $\lambda_{7}$ to be $1.0$, $0.5$, $1.0$, $0.5$, $0.5$, $0.1$ and $0.5$, respectively. For training, a 512$\times$1024 image patch with the min-batch size of 4 is used to train the network. It is noteworthy that we only save the generator $G_{s}$ model for testing.

\begin{table}[!t] 
\footnotesize
\setlength{\tabcolsep}{1.0mm}
	\caption{Quantitative evaluation of different rain removal and haze removal methods on the testing sets of RainCityscapes and Rain200H. Note that we train all semi-supervised methods on RainCityscapes$\&$MOR-Rain200 and Rain200H$\&$MOR-Rain200. And all supervised methods are directly trained on paired data of RainCityscapes and Rain200H.} 	
	\centering
	\begin{tabular}{ l|c|c c||c c }
		\hline
		   \multirow{2}{*}{Methods} & \multirow{2}{*}{Publication} &\multicolumn{2}{c||}{RainCityscapes} &\multicolumn{2}{c}{Rain200H} \\
		    \cline{3-6}
		        & & PSNR  & SSIM  & PSNR  & SSIM  \\\hline
		\multicolumn{6}{c}{Rain Streaks Removal} \\\hline
		DSC & ICCV'2015 & 16.41 & 0.771 & 15.29 & 0.423\\ 
		GMMLP & CVPR'2016 & 18.39 & 0.819 & 14.54 & 0.548\\ 
		JOB & ICCV'2017 & 14.79 & 0.752 & 16.21 & 0.523\\ 
		DID-MDN &  CVPR'2018 & 22.31 & 0.817 & 24.49 & 0.843 \\
		RESCAN &  ECCV'2018 & 24.30 & 0.842 & 26.64 & 0.845 \\ 
		UMRL &  CVPR'2019 & 27.97 & 0.912 & 27.27& 0.898\\ 
		PReNet &  CVPR'2019 & 26.83 & 0.910 & 28.08 & 0.887\\ 
		SPANet &  CVPR'2019 & 31.56 & 0.931 & 23.85 & 0.852 \\ 
		SIRR &  CVPR'2019 & 23.14 & 0.832  & 25.25 & 0.863 \\
		DCSFN &  ACM MM'2020 & 26.37 & 0.872 & 28.26 & 0.899\\
		MSPFN &  CVPR'2020 & 26.89 & 0.903 & 27.21 & 0.896 \\
		Syn2Real &  CVPR'2020 & 28.66 & 0.919 & 26.37 & 0.874 \\
		DerainRLNet & CVPR'2021 & 27.39 & 0.881 & \underline{28.87} & \underline{0.895}\\
		MPRNet  &  CVPR'2021 & 25.76 & 0.857 & 28.39 & 0.887 \\
		JRGR &  CVPR'2021 & 22.39 & 0.821 & 26.12 & 0.873 \\ \hline
        \multicolumn{6}{c}{Rainy Haze Removal} \\\hline
		 DCPDN & CVPR'2018 & 28.45 & 0.896 & 19.72 & 0.710\\ 
		 EPDN & CVPR'2019 & 25.87 & 0.905 & 19.59 & 0.731 \\ 
		 FFA & AAAI'2020 & 28.97 & 0.884 & 20.12 & 0.762\\\hline 
		\multicolumn{6}{c}{Rain Streaks \& Rainy Haze Removal} \\\hline
		DAF-Net &  CVPR'2019 & 30.66 & 0.924 & 24.65 & 0.860\\ 
		DGNL-Net &  TIP'2021 & \underline{32.21} & \underline{0.936}  & 27.79 & 0.886 \\ 
	    \textbf{Ours}&submit'22&\textbf{35.67} & \textbf{0.948} & \textbf{29.58} & \textbf{0.905} \\
		\hline
	\end{tabular} \label{Quantitative comparisons}
\end{table}

\subsection{Evaluation Settings and Datasets}
We qualitatively and quantitatively compare the performance of our Semi-MoreGAN with different types of rain removal and haze removal approaches: 1) semi-supervised rain removal methods include Syn2Real~\cite{yasarla2020syn2real}, SIRR~\cite{wei2019semi} and JRGR \cite{ye2021closing}; 2) supervised rain removal methods include DCSFN~\cite{wang2020dcsfn}, SPANet~\cite{wang2019spatial}, PReNet~\cite{ren2019progressive}, UMRL~\cite{yasarla2019uncertainty}, DID-MDN~\cite{zhang2018density}, RESCAN~\cite{li2018recurrent}, JOB~\cite{zhu2017joint}, GMMLP~\cite{li2016rain}, DSC~\cite{luo2015removing}, MSPFN \cite{jiang2020multi}, DerainRLNet \cite{chen2021robust} and MPRNet \cite{zamir2021multi}; 3) image haze removal methods include FFA~\cite{qin2020ffa}, EPDN~\cite{qu2019enhanced} and DCPDN~\cite{zhang2018densely}. Moreover, we also compare our results with DGNL-Net~\cite{hu2021single} and DAF-Net~\cite{hu2019depth} which can remove both rain streaks and rainy haze together.

As Semi-MoreGAN is trained in a semi-supervised learning manner, both synthetic and real-world rainy images are utilized for training our model. The synthetic datasets are: (1) Rain200H \cite{yang2017deep} containing 1800 training images and 200 testing images with five streak directions; (2) RainCityscapes dataset \cite{hu2019depth} containing 9432 training images and 1188 testing images with rain streaks and rainy haze. For the real-world dataset, we collect a total of 1100 unpaired real-world rainy images from the datasets provided by \cite{wei2019semi,zhang2019image,yang2017deep,wu2019beyond,li2019single} and Google search. Then, we randomly collect 300 real-world rainy images (200 training images and 100 testing images) to build a new real-world rainy dataset named MOR-Rain200. Following the protocols of \cite{yasarla2020syn2real}, we train all semi-supervised methods using synthetic datasets (Rain200H and RainCityscapes) as paired data and MOR-Rain200 as unpaired data, which are denoted by $\&$, such as RainCityscapes$\&$MOR-Rain200 and Rain200H$\&$MOR-Rain200. And all supervised methods are directly trained on paired data of RainCityscapes and Rain200H.

\begin{table}[!t]
\footnotesize
\centering
\caption{Quantitative evaluation of the combinations of rain removal and haze removal methods which achieve relatively good performance on the testing set of RainCityscapes.} 
\begin{tabular}{c|c|c}
\hline
Method                               & PSNR                       & SSIM                        \\ \hline
Syn2Real+FFA                         & 27.79                      & 0.872                   \\ 
Syn2Real+DCPDN                       & 24.19                      & 0.839                    \\ 
MPRNet+FFA                             & 29.73                      & 0.930                     \\ 
MPRNet+DCPDN                           & 27.11                      & 0.844                     \\ \hline
\end{tabular}

\label{combine}
\end{table}
\begin{figure*}[ht]
	\centering
	\includegraphics[width=1.0\linewidth]{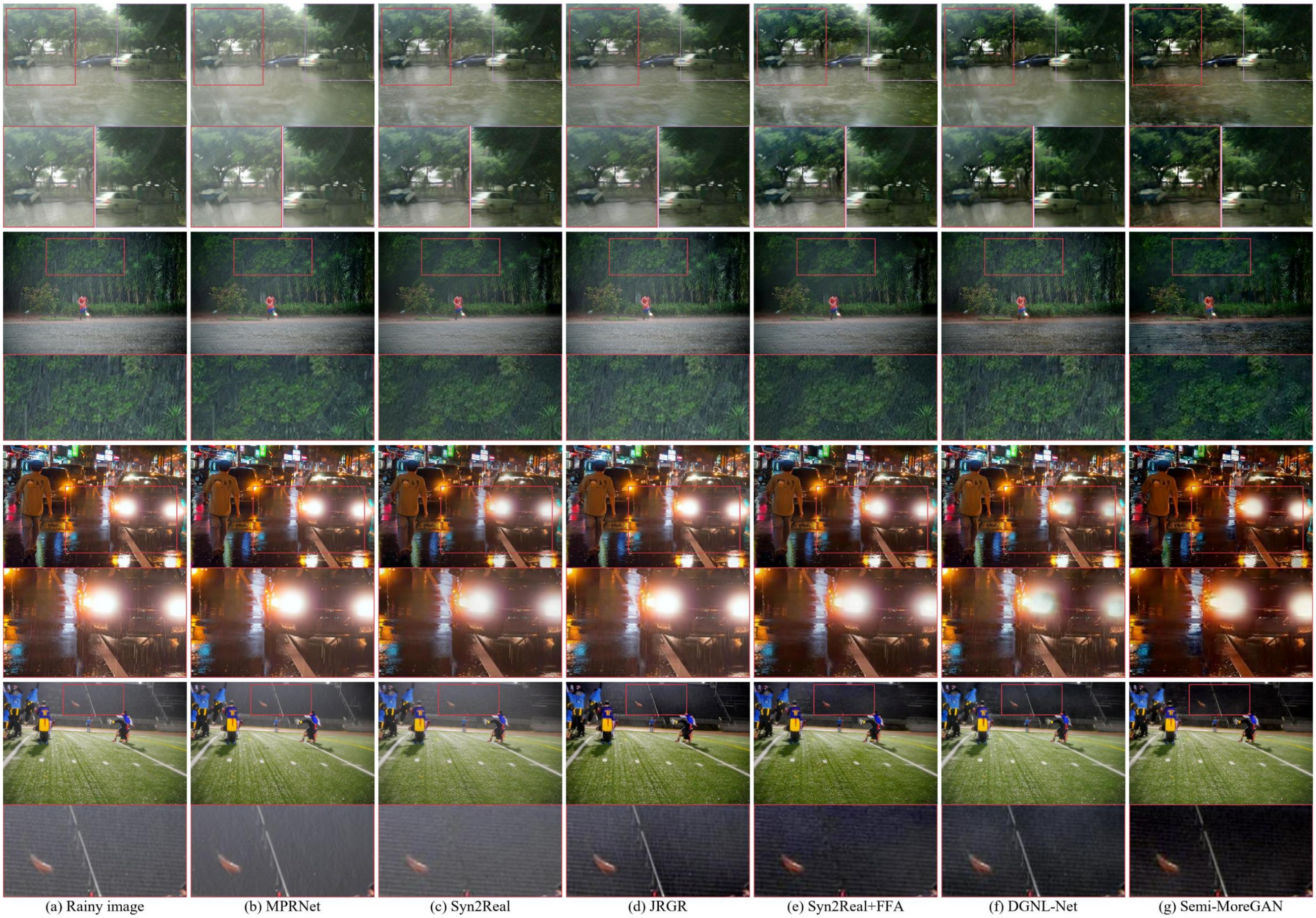}
	\caption{Visual comparison results on heavy rainy images. Note that all models are trained on RainCityscapes$\&$MOR-Rain200.} 
	\label{fig:real_heavy}
\end{figure*}
\subsection{Comparison on Synthetic Dataset}
\par We compare our method with several state-of-the-art rain removal and haze removal approaches on the testing set of RainCityscapes and Rain200H. Since Rain200H does not provide depth maps, we follow the protocol of \cite{hu2019depth,hu2021single} to assume a constant depth value of 0.5 on the whole image. This means that we simply ignore the depth and add rain streaks as a 2D overlay on the rain-free input photos. We adopt this strategy to train the models of DGNL-Net~\cite{hu2021single}, DAF-Net~\cite{hu2019depth} and Semi-MoreGAN.

The quantitative results are illustrated in Table~\ref{Quantitative comparisons}, the proposed method ranks the first among all methods. Especially, our method obtains almost 3.46dB and 1.79dB PSNR gains on RainCityscapes and Rain200H, compared to DGNL-Net~\cite{hu2021single}, which originally builds the RainCityscapes dataset for mixture of rain removal. 
Besides, Semi-MoreGAN still outperforms all semi-supervised rain removal methods. For example, Semi-MoreGAN obtains more than 13.28dB and 7.01dB PSNR
gains on RainCityscapes compared with JRGR \cite{ye2021closing} and Syn2Real~\cite{yasarla2020syn2real}. Except for heavy rain cases, our method is also capable of synthesizing normal rainy images. For instance, compared to the latest supervised method DerainRLNet \cite{chen2021robust}, our semi-supervised paradigm obtains more than 0.71dB and 0.10 gains in PSNR and SSIM on Rain200H respectively.

\par The qualitative results are shown in Fig. \ref{fig:f1}. It is observed that the rain removal results of Syn2Real~\cite{yasarla2020syn2real}, MPRNet \cite{zamir2021multi} and Syn2Real~\cite{yasarla2020syn2real}+FFA~\cite{qin2020ffa} remain some rain streaks and rainy haze and fail to preserve the original color maps. DGNL-Net~\cite{hu2021single} can remove large rain streaks but remain some small rain streaks and rainy haze. In contrast, our Semi-MoreGAN evidently alleviates these deficiencies and obtains the closest result to the ground truth.

\par Observing that most rain removal methods fail to remove the rainy haze followed by rain streaks, thus we also combine several rain removal and haze removal approaches that achieve good performance in Table \ref{Quantitative comparisons}, but they still fail to derive better performance in mixture of rain removal (shown in Table \ref{combine}). It is illustrated that simple cascaded combinations ignore the essential depth maps, thus limiting the performance on mixture of rainy images, and even losing more details (Fig. \ref{fig:f1} (d)).


\subsection{Comparison on Real-world Rainy Images}
We also evaluate the proposed method on real-world images from the testing set of MOR-Rain200, Fig. \ref{fig:real_heavy} show the results on real-world rainy images. For heavy rain removal, JRGR \cite{ye2021closing} and MPRNet \cite{zamir2021multi} neither effectively remove rain streaks nor rainy haze. Syn2Real~\cite{yasarla2020syn2real} and DGNL-Net~\cite{hu2021single} can remove part of small rain streaks but fail for large rain streaks and the rainy haze.
Syn2Real~\cite{yasarla2020syn2real} even brings details lost (the white pillars in Row 4). Besides, the results of Syn2Real~\cite{yasarla2020syn2real}+FFA~\cite{qin2020ffa} remain the majority of rain, and the haze removal method may darken the picture. 
In contrast, our method can well handle the rain streaks and rainy haze while preserving structure details under heavy rain conditions. 
More qualitative results on real-world rainy images are provided in the supplementary material.
%


\subsection{Ablation Study} \label{ablation}
\textbf{Component Analysis.} We evaluate the effectiveness of different components of Semi-MoreGAN as follows: 
\begin{itemize}
\item[$\bullet$]  \textbf{BL:} Baseline (BL) indicates contextual feature prediction network without 
contextual feature aggregation blocks (only consists of four convolution layers and three non-linear ReLU layers), which learns a function that maps the rainy images to the rain-free images.
\item[$\bullet$]  \textbf{CFPN:} Adding contextual feature aggregation blocks to the baseline and comprising the whole contextual feature prediction network (CFPN).
\item[$\bullet$]  \textbf{CFPN+DPN:} Employing two sub-networks for depth estimation and image rain removal. One network is the contextual feature prediction network (CFPN), and the other is the depth prediction network (DPN), which denotes ADPN without a self-attention block \cite{vaswani2017attention}. Note that this model integrates contextual features with depth maps via a common multiplication operation.
\item[$\bullet$]  \textbf{CFPN+ADPN:} Adding a self-attention block \cite{vaswani2017attention} to the depth prediction network and comprising the whole attentional depth prediction network (ADPN).
\item[$\bullet$]  \textbf{CFPN+ADPN+NLN:} Comprising the contextual feature prediction network, the attentional depth prediction network, and the standard non-local network (NLN), which denotes PDNL without pyramid pooling layers and additional attention module.
\item[$\bullet$]  \textbf{CFPN+ADPN+PDNL:} Our Semi-MoreGAN, which comprises the contextual feature prediction network, the attentional depth prediction network, and the pyramid depth-guided non-local network.
\end{itemize}
As shown in Table \ref{fig:ablation1}, the full structure of Semi-MoreGAN achieves the highest performance in terms of PSNR and SSIM, which indicates that all the components of Semi-MoreGAN are beneficial for effective mixture of rain removal.
Besides, we also conduct experiments to adapt different number settings of CFABs and compare DGNL\cite{hu2021single} with our PDNL, which are provided in the supplementary material.
%



\begin{table}[!t]
\footnotesize
\setlength{\tabcolsep}{1.0mm}
\centering
\caption{Rain removal performance (PSNR/SSIM) on the testing set of RainCityscapes with different components of Semi-MoreGAN. Note that M-A, M-B, M-C, M-D, M-E and Ours denotes BL, CFPN, CFPN+DPN, CFPN+ADPN, CFPN+ADPN+NLN and CFPN+ADPN+PDNL, respectively.}
\begin{tabular}{cccccccc}
\toprule
Models&M-A& M-B& M-C& M-D&M-E&Ours \\ \midrule
PSNR     &21.33&27.19&29.46&31.97&33.62&\textbf{35.67} \\ 
SSIM     &0.801&0.872&0.919&0.928&0.937&\textbf{0.948} \\ \bottomrule
\end{tabular}
\label{fig:ablation1}
\end{table}
\begin{table}[!t]\footnotesize
\setlength{\tabcolsep}{1.0mm}
\centering
\caption{Deraining performance (PSNR/SSIM) on the testing set of RainCityscapes with different loss functions of Semi-MoreGAN. Note that w/o denotes without, $L_{adv}$ and $L_{adv'}$ denote $L_{adv-super}$ and $L_{adv-unsuper}$, respectively.}
\begin{tabular}{ccccccccc}
\toprule
$Variants$ &$V_0$& $V_1$ & $V_2$ &$V_3$ & $V_4$&$V_5$&$V_6$&$V_7$ \\\midrule
$L_{multi}$&  w/o&  \checkmark  & \checkmark   &  \checkmark  &  \checkmark& \checkmark& \checkmark & \checkmark\\
$L_{adv}$&  w/o&  w/o  & \checkmark   &  \checkmark  &  \checkmark & \checkmark& \checkmark& \checkmark \\
$L_{cyc}$&  w/o&   w/o  &  w/o   &  \checkmark  &  \checkmark& \checkmark& \checkmark& \checkmark\\
$L_{adv'}$&  w/o&   w/o  &  w/o   & w/o    & \checkmark& \checkmark& \checkmark& \checkmark \\ 
$L_{dc}$&  w/o&   w/o  &  w/o   & w/o    & w/o & \checkmark& \checkmark & \checkmark\\ 
$L_{tv}$&  w/o&   w/o  &  w/o   & w/o    & w/o & w/o & \checkmark & \checkmark\\ 
$L_{per}$&  w/o&   w/o  &  w/o   & w/o    & w/o & w/o & w/o & \checkmark\\ 
\hline
PSNR&  31.13     &  33.11  & 33.79 & 34.03    &34.36  & 34.69& 34.78& 35.67\\
SSIM&  0.926     &  0.936  & 0.938    &0.940    &0.941  & 0.943&0.943& 0.948 \\\bottomrule
\end{tabular}
\label{fig:ablation2}
\end{table}
\textbf{Loss Function Analysis.} Apart from the analysis of different components of Semi-MoreGAN, we explore the contribution of different loss functions in our model: 
\begin{itemize}
\item[$\bullet$]  only rain reconstruction loss $\parallel Gs(x_{r})-x_g\parallel_1$ $\rightarrow$ $V_0$.
\item[$\bullet$]  only multi-task loss $L_{multi}$ $\rightarrow$ $V_1$.
\item[$\bullet$]  $V_1$ + supervised adversarial loss $L_{adv-super}$ $\rightarrow$ $V_2$.
\item[$\bullet$]  $V_2$ + cycle-consistency Loss $L_{cyc}$ $\rightarrow$ $V_3$.
\item[$\bullet$]  $V_3$ + unsupervised adversarial loss $L_{adv-unsuper}$ $\rightarrow$ $V_4$.
\item[$\bullet$]  $V_4$ + dark channel loss $L_{dc}$ $\rightarrow$ $V_5$.
\item[$\bullet$]  $V_5$ + TV loss $L_{tv}$ $\rightarrow$ $V_6$.
\item[$\bullet$]  $V_6$ + perceptual loss $L_{per}$ $\rightarrow$ $V_7$ (full model).
\end{itemize}
as the discriminators are only compatible with their losses, thus without adversarial loss also means without paired discriminator. Besides, $V_{0-2}$ models only contain supervised branches and are trained on RainCityscapes. $V_{3-7}$ models are semi-supervised models which are trained on
RainCityscapes$\&$MOR-Rain200. Table \ref{fig:ablation2} shows that the performance of Semi-MoreGAN gradually improves when the loss functions are progressively incorporated. Especially, the $V_1$ model obtains more than 1.98dB PSNR gains on RainCityscapes compared with $V_0$, which verifies the effectiveness of the joint learning paradigm of mixture of rain removal and depth estimation. The supervised version of Semi-MoreGAN ($V_{0-2}$) has achieved comparable performance against other supervised approaches, and the semi-supervised models ($V_{3-7}$) can take advantage of the unlabeled real-world dataset MOR-Rain200 to obtain better rain removal performance with the help of a variety of unsupervised loss functions.


\textbf{Data Paradigm Analysis.}
Since Semi-MoreGAN is trained in a semi-supervised manner, we also analyze the capacity of Semi-MoreGAN to use different data paradigms for training the model on RainCityscapes and real-world rainy images, which are provided in the supplementary material.

\begin{table*}[!t]\footnotesize
\setlength{\tabcolsep}{1.0 mm}
\centering
\caption{Comparisons on runtime (seconds) and GFlops.}
\begin{tabular}{cccccccccccccccc}
\toprule
Method  & UMRL & PReNet & SPANet&SIRR&DCSFN&MSPFN&Syn2Real&DerainRLNet&MPRNet&JRGR&DAF-Net&DGNL-Net&Semi-MoreGAN\\ \midrule
Time&1.35s&0.26s&0.20s&0.32s&1.52s&3.86s&1.84s&1.49s&3.67s&1.76s&1.93s&0.76s&0.48s\\ 
GFlops&119.57&88.76&149.13&35.32&594.88&702.78&327.44&548.92&989.64&129.6&168.61&119.54&58.9 \\ \bottomrule
\end{tabular}
\label{Tab:time}
\end{table*}
\subsection{Impact of Precise Depth Prediction} \label{depth_ab}
We conduct experiments to explore the impact of precise depth prediction, please refer to the supplementary material.

\subsection{Application}
To provide further evidence that the visibility enhancement of Semi-MoreGAN could be helpful for computer vision applications, we employ Google Vision API to evaluate our rain removal results, which are provided in the supplementary material.


\subsection{Running Time}
\par We compare the running time and GFlops of our method with different approaches as shown in Table \ref{Tab:time}. We use the 512$\times$1024 image patch for evaluation. It is observed that our method is not the fastest one, but its performance is still acceptable.


\section{Conclusion}
We explore the visual effects of mixture of rain streaks and rainy haze and propose Semi-MoreGAN. Semi-MoreGAN leverages both synthetic datasets and real-world rainy images for training, thus generalizing smoothly in real-world scenarios. Considering the visual effects of rain subject to scene depth, we propose a novel attentional depth prediction network (ADPN) to obtain more robust and sharper depth maps, thus boosting the mixture of rain removal ability of the model. Besides, a contextual feature prediction network is proposed to produce image features with more context information. Additionally, we propose a pyramid depth-guided non-local network  to effectively integrate the image features with the depth information, and produce the final rain-free images. Comprehensive evaluations show that Semi-MoreGAN outperforms existing rain removal models, on both synthetic and real rain data. In the future, we will further explore the potential of Semi-MoreGAN for mixture degradation problems caused by photographing in various scenarios under bad weather and nighttime conditions, and extend Semi-MoreGAN to the application of real-time video by exploiting the temporal information in the video.

\section{Acknowledgments}
This work was supported by the National Key Research and Development Program of China (No. 2021ZD0113200, No. 2020YFB1805400), the National Natural Science Foundation of China (No. 42071431, No. 62172218), the Provincial Key Research and Development Program of Hubei, China (No. 2020BAB101),  the Direct Grant (No. DR22A2) and the Research Grant entitled "Self-Supervised Learning for Medical Images" (No. 871228) of Lingnan University, Hong Kong.

%
%












\bibliographystyle{eg-alpha-doi}  
\bibliography{refer,dehaze,dl_streak,prior}        


\end{document}